\documentclass{article}

\usepackage{microtype}
\usepackage{graphicx}
\usepackage{subcaption}
\usepackage{booktabs} % for professional tables

\usepackage{hyperref}
\usepackage{pifont}
\usepackage{amsmath,amsfonts,bm}

\def\eqref#1{equation~\ref{#1}}
\def\1{\bm{1}}

\def\vc{{\bm{c}}}

\def\vl{{\bm{l}}}

\def\vs{{\bm{s}}}

\def\vv{{\bm{v}}}

\def\mC{{\bm{C}}}

\def\mL{{\bm{L}}}

\def\mP{{\bm{P}}}

\def\mS{{\bm{S}}}

\def\mU{{\bm{U}}}

\def\mW{{\bm{W}}}

\def\mZ{{\bm{Z}}}

\DeclareMathAlphabet{\mathsfit}{\encodingdefault}{\sfdefault}{m}{sl}
\SetMathAlphabet{\mathsfit}{bold}{\encodingdefault}{\sfdefault}{bx}{n}

\usepackage[preprint]{icml2026}

\usepackage{amsmath}
\usepackage{amssymb}
\usepackage{mathtools}
\usepackage{amsthm}

\usepackage[capitalize,noabbrev]{cleveref}

\theoremstyle{plain}

\theoremstyle{definition}

\theoremstyle{remark}

\usepackage[textsize=tiny]{todonotes}

\icmltitlerunning{Learning Sparse Visual Representations via Spatial-Semantic Factorization}

\begin{document}

\twocolumn[
  \icmltitle{Learning Sparse Visual Representations via Spatial-Semantic Factorization}

  % It is OKAY to include author information, even for blind submissions: the
  % style file will automatically remove it for you unless you've provided
  % the [accepted] option to the icml2026 package.

  % List of affiliations: The first argument should be a (short) identifier you
  % will use later to specify author affiliations Academic affiliations
  % should list Department, University, City, Region, Country Industry
  % affiliations should list Company, City, Region, Country

  % You can specify symbols, otherwise they are numbered in order. Ideally, you
  % should not use this facility. Affiliations will be numbered in order of
  % appearance and this is the preferred way.
  \icmlsetsymbol{equal}{*}

  \begin{icmlauthorlist}
    \icmlauthor{Theodore Zhengde Zhao}{ms}
    \icmlauthor{Sid Kiblawi}{ms}
    \icmlauthor{Jianwei Yang}{x,done}
    \icmlauthor{Naoto Usuyama}{ms}
    \icmlauthor{Reuben Tan}{ms}
    \icmlauthor{Noel C Codella}{ms}
    \icmlauthor{Tristan Naumann}{ms}
    %\icmlauthor{}{sch}
    \icmlauthor{Hoifung Poon}{ms}
    \icmlauthor{Mu Wei}{ms}
    %\icmlauthor{}{sch}
    %\icmlauthor{}{sch}
  \end{icmlauthorlist}

  \icmlaffiliation{ms}{Microsoft}
  \icmlaffiliation{done}{Work done at Microsoft}
  \icmlaffiliation{x}{xAI}

  \icmlcorrespondingauthor{Theodore Zhengde Zhao}{theodorezhao@microsoft.com}
  \icmlcorrespondingauthor{Mu Wei}{muwei@microsoft.com}

  % You may provide any keywords that you find helpful for describing your
  % paper; these are used to populate the "keywords" metadata in the PDF but
  % will not be shown in the document
  \icmlkeywords{Machine Learning, ICML}

  \vskip 0.3in
]

% this must go after the closing bracket ] following \twocolumn[ ...

% This command actually creates the footnote in the first column listing the
% affiliations and the copyright notice. The command takes one argument, which
% is text to display at the start of the footnote. The \icmlEqualContribution
% command is standard text for equal contribution. Remove it (just {}) if you
% do not need this facility.

% Use ONE of the following lines. DO NOT remove the command.
% If you have no special notice, KEEP empty braces:
\printAffiliationsAndNotice{}  % no special notice (required even if empty)
% Or, if applicable, use the standard equal contribution text:
% \printAffiliationsAndNotice{\icmlEqualContribution}

\begin{abstract}
Self-supervised learning (SSL) faces a fundamental conflict between semantic understanding and image reconstruction. High-level semantic SSL (e.g., DINO) relies on global tokens that are forced to be location-invariant for augmentation alignment, a process that inherently discards the spatial coordinates required for reconstruction. Conversely, generative SSL (e.g., MAE) preserves dense feature grids for reconstruction but fails to produce high-level abstractions. We introduce STELLAR, a framework that resolves this tension by factorizing visual features into a low-rank product of semantic concepts and their spatial distributions. This disentanglement allows us to perform DINO-style augmentation alignment on the semantic tokens while maintaining the precise spatial mapping in the localization matrix necessary for pixel-level reconstruction. We demonstrate that as few as 16 sparse tokens under this factorized form are sufficient to simultaneously support high-quality reconstruction (2.60 FID) and match the semantic performance of dense backbones (79.10\% ImageNet accuracy). Our results highlight STELLAR as a versatile sparse representation that bridges the gap between discriminative and generative vision by strategically separating semantic identity from spatial geometry. Code available at \url{https://aka.ms/stellar}.

% Self-supervised vision encoders have become critical components of modern machine learning systems. Despite remarkable advances in image understanding, generation, and multimodal alignment, the underlying representation of visual features has remained largely unchanged, constrained by historical architectures and benchmarks. This reliance on dense feature grids introduces redundancy and limits the integration of understanding and generation. We propose a novel framework that represents images with a small number of sparse tokens \rev{in the form of} low-rank matrix factorization. While mathematically simple, this formulation effectively disentangles semantic and spatial information. We demonstrate that vision-only self-supervised learning under this framework yields sparse token representations that simultaneously support high-quality image understanding, detailed pixel-level reconstruction, and fine-grained semantic understanding. Together, these results highlight sparse tokens as a promising alternative to dense grids for efficient and versatile visual representation learning.

\end{abstract}
\section{Introduction}

Learning visual representations has been a central pursuit in computer vision since the advent of deep learning~\citep{bengio2013representation}. Modern vision models encode raw pixels into latent features powering nearly all downstream applications. Despite advances from early convolutional networks~\citep{CNN} to ResNets~\citep{he2016deep} and vision transformers (ViTs)~\citep{dosovitskiy2020image}, the geometric format of visual representation has remained largely unchanged: a dense 2D grid of high-dimensional features, where each vector is tied to a local patch. This design is intuitive, as it mirrors the grid-like arrangement of pixels.

On the other hand, the field faces a longstanding dilemma: the pursuit of a unified, \textit{holistic} representation that excels at both high-level semantic understanding and low-level reconstruction. While this synthesis has succeeded in natural language processing, where reconstruction tasks like bidirectional masking (BERT~\cite{devlin2019bert}) or autoregressive modeling (GPT~\cite{brown2020language}) naturally induce superior semantics, it doesn't directly transfer to the vision domain. Representations learned primarily through image reconstruction (e.g., MAE~\citep{he2022masked}, SimMIM~\cite{xie2022simmim}) often yield semantics that trail behind contemporary state-of-the-art methods. Consequently, recent self-supervised learning (SSL) approaches have largely diverged into two camps: those prioritizing pixel-level grounding via reconstruction, and those prioritizing rich semantics via joint-embedding invariance~\cite{van2025joint}.

We argue that this divergence stems from an \textit{Invariance Paradox} inherent to the dense grid format. For a dense representation to faithfully reconstruct an image, it must preserve precise spatial information which are inherently \textit{equivariant} to transformations like cropping or shifting. Conversely, high-level semantics are \textit{invariant} to such transformations. Traditional SSL methods such as DINO~\cite{caron2021emerging} attempt to force invariance onto global representations obtained from these dense grids. This creates a fundamental conflict: the model is pressured to discard spatial variance to achieve semantic alignment, yet the dense grid format requires equivariance for spatial grounding.

In this work, we show that this paradox is not an inevitable trade-off, but a byproduct of the dense representation itself. By moving away from the 2D grid towards a \textit{sparse, factorized latent representation}, we can jointly achieve high-fidelity reconstruction and rich semantics. Our key insight is that the information necessary to describe a scene can be disentangled into two complementary, sparse factors:
\begin{enumerate}
    \item \textbf{The ``What''}: A set of sparse latent tokens representing invariant visual concepts.
    \item \textbf{The ``Where''}: A set of equivariant coefficients representing their spatial locations.
\end{enumerate}

By disentangling these factors through a low-rank matrix factorization form, we enable a ``semantic triage'': the model is forced to reconstruct the entire image using a highly compressed bottleneck. This encourages the model to ignore stochastically redundant background pixels and focus on semantically-rich object regions. We propose \textbf{STELLAR}, a framework that achieves high-quality reconstruction from as few as 16 tokens while encoding fine-grained semantics in a fully self-supervised manner.

Our contributions are summarized as follows:
\begin{itemize}
    \item \textbf{Sparse Representation}: We propose STELLAR, an efficient form of vision modeling that factorizes an image into a handful of sparse tokens by disentangling \textit{what} concepts are present from \textit{where} they are located.
    \item \textbf{SSL Method}: We introduce a training scheme to learn these representations without annotation. By aligning visual concepts across views using optimal transport, we enforce invariance in the ``what'' factor while adapting the ``where'' factor, inducing rich semantics.
    \item \textbf{Empirical Observations}: (i) STELLAR achieves a state-of-the-art balance of semantics (IN-1K linear acc. 79.10\%) and reconstruction (FID 2.60), outperforming prior approaches. (ii) Our sparse image modeling induces fine-grained, region-aware semantics even without explicit dense supervision, outperforming prior work with similar training budget.
\end{itemize}

\section{Related Work}

\textbf{Self-supervised Learning.} Modern SSL generally falls into two paradigms. \textit{Joint Embedding} (JE) methods, such as the MoCo~\citep{he2020momentum} and DINO~\citep{oquab2023dinov2} families, prioritize global invariance via multi-view alignment, yielding strong semantics but often losing spatial grounding. Conversely, \textit{Masked Image Modeling} (MIM), exemplified by MAE~\citep{he2022masked} and SimMIM~\citep{xie2022simmim}, emphasizes spatial equivariance through pixel reconstruction. While hybrids like iBOT~\citep{zhou2021ibot} and DINOv2~\citep{oquab2023dinov2} attempt to combine these objectives, they still rely heavily on global invariance and forgo pixel reconstruction. 

\textbf{Sparse Representation.} A growing body of work replaces dense feature maps with compact embeddings. Sparse R-CNN~\citep{sun2021sparse} and Mask2Former~\citep{cheng2022masked} utilize sparse queries for supervised tasks, while BLIP-2~\citep{li2023blip} and TiTok~\citep{yu2024image} employ sparse tokens for vision--language or generative efficiency. SemMAE~\citep{li2022semmae} utilizes sparse tokens to guide masking using a pretrained teacher. Unlike these methods, STELLAR treats sparse tokens as the \textbf{primary latent representation} and learns in SSL manner.

\textbf{Disentanglement \& Low-rank Factorization.} The assumption that high-dimensional data lie on low-dimensional manifolds is foundational to dictionary learning~\citep{mairal2008supervised}. In deep learning, low-rank constraints are typically applied to weights for efficiency (e.g., LoRA~\citep{hu2022lora}). STELLAR differs by applying \textit{low-rank factorization to the feature map itself}, disentangling "what" (semantic latents $\mL$) from "where" (spatial assignments $\mS$).

\paragraph{The Empirical Dilemma.} Current vision frameworks face a persistent gap: models excelling at pixel-level reconstruction often produce weaker semantic representations~\citep{zhang2022improving, chen2024deconstructing}, while those achieving top-tier semantics often abandon reconstruction to avoid low-level shortcuts~\citep{assran2023self, darcet2025cluster}. We demonstrate that by factorizing the latent representation, it is possible to achieve strong performance on both image understanding and reconstruction.

\section{Preliminaries}
Representation learning involves encoding an image $X \in \mathcal{X}$ to latent features $\mZ(X)$ for downstream tasks. Traditionally, vision representations take a \textit{dense} spatial form:
\[
\mZ \in \mathbb{R}^{n \times d},
\]
where $n = h \times w$ denotes the number of patches on a dense grid that partitions the image. Each grid location is represented by a feature vector $\mathbf{z}_i := \mZ_{i,:} \in \mathbb{R}^d$ for $1 \leq i \leq n$. Most vision architectures also incorporate a global representation $\mathbf{z}_0 \in \mathbb{R}^d$, typically obtained via global pooling or a specialized \texttt{[CLS]} token that undergoes self-attention with patch tokens.

Ideally, we want $\mZ$ to serve as a \textit{holistic} representation of the image $X$, which retains sufficient information about the image details, while at the same time possesses rich semantics for downstream tasks. Mathematically, we define such representation as follows:

\begin{itemize}
    \item \textbf{Reconstruction}: There exists a decoder $\mathcal{D}$ such that $\mathcal{D}(\mathbf{Z}(X)) \approx X$. This ensures the representation is spatially and texturally grounded in the physical input.
    \item \textbf{Semantics}: For a downstream task with joint distribution $(X, Y) \sim \mathcal{X} \times \mathcal{Y}$, there exists a simple predictor $f \in \mathcal{F}$ (e.g., a linear layer) such that the expected task loss $\mathbb{E}_{(X,Y)} \big[ \mathcal{L}(f(\mathbf{Z}(X)), Y) \big]$ is minimized using frozen features. Typically $Y$ reflects human perception.
\end{itemize}

Current SSL paradigms are caught in a fundamental ``\textit{Invariance Paradox}''. In order to learn high-level semantics, Joint Embedding (JE) methods (e.g. the DINO family) impose \textit{invariance} to spatial transformations, even when the image is cropped to as small as only 5\%.  On the other hand, reconstruction requires spatial detail, because every pixel shift requires a different set of features for precise reconstruction. This results in representations which are highly \textit{equivariant} to the transformation, i.e. the feature map transforms along with the transformation in the image.

Let $\mathcal{T}$ be a group of spatial transformations (e.g., translations), and $t_\theta \in \mathcal{T}$ is parametrized by $\theta$ i. A representation $\mZ (X)$ suffers from the \textit{Invariance Paradox} if it must simultaneously satisfy two contradictory constraints: 
\begin{itemize}
    \item \textbf{Semantic Invariance:} The representation should be insensitive to $t_\theta \in \mathcal{T}$: 
    \[
    \|\frac{\partial}{\partial \theta} \mZ (t_\theta \circ X)\|_F \approx 0.
    \]
    \item \textbf{Spatial Equivariance:} To allow for high-fidelity reconstruction, the representation must track spatial shifts: $\mathcal{D}(\mZ (t_\theta \circ X)) \approx t_\theta \circ X$. With chain rule and matrix norm inequalities, we have
    \[
    \|\frac{\partial}{\partial \theta}\mZ (t_\theta \circ X)\|_F \gtrsim \frac{\left\| \frac{\partial (t_\theta \circ X)}{\partial \theta} \right\|_F}{\sigma_{\max} \left( \frac{\partial \mathcal{D}}{\partial \mZ } \right)} > 0.
    \]
\end{itemize}

% In contrast, \textit{sparse visual representation} aims to represent the image with
% \[
% \mZ \in \mathbb{R}^{r \times d}, \quad r \ll h \cdot w.
% \]

% Ideally, the number of sparse tokens $r$ should be less than an order of magnitude than the total number of dense tokens $n = h\cdot w$. 
% Prior works tend to emphasize only one aspect of “representation.” For example, TiTok~\citep{yu2024image} uses 32 tokens to reconstruct an image with 256 patches at high fidelity, but its sparse features lag far behind other self-supervised vision models in semantic understanding. Conversely, MAE~\citep{he2022masked} is designed to capture semantics by reconstructing randomly masked patches, yet the resulting reconstructions are often blurry. 

\section{The STELLAR Framework}
\subsection{Sparse Image Modeling}

From now on we consider a form of representation in alternative to the dense grid-based representations describing what appears at each individual location. We start from the principle that an image depicts the physical world, which can be understood as a collection of objects located in space. 

To begin with, we model an image with a compact set of semantic concepts together with their spatial distributions. Let there be $r$ concept embeddings $\vs_1, \cdots, \vs_r \in \mathbb{R}^d$,
where each $\vs_j$ captures a distinct semantic concept. The spatial distribution of these concepts is expressed through weights $\vl_1, \cdots, \vl_n \in \mathbb{R}^{r}$, where $n$ is the total number of patches. 

By constraining $0 \leq \vl_i \leq 1$ and $\mathbf{1}^\top \vl_i = 1$, each patch is represented as a convex combination of the concept embeddings: $\vv_i = \sum_{j=1}^r \vl_{i,j} \vs_j$. Thus, the set ${\vs_j}_{j=1}^r$ acts as a basis for constructing local features. In matrix form, the latent representation now takes the form
\begin{equation}\label{eq-lowrank}
    \mZ(X) = \mL(X) \mS(X), 
\end{equation}

where
$\mS = [\vs_1, \ldots, \vs_r]^\top \in \mathbb{R}^{r \times d}$ is the \textit{semantic matrix}, and $\mL = [\vl_1, \ldots, \vl_n]^\top \in \mathbb{R}^{n \times r}$ is the \textit{localization matrix}, with the constraint $0 \leq \mL \leq 1, \mL \mathbf{1}_r =\mathbf{1}_n$.

Compared to a canonical dense representation of shape $n\times d$, $\mZ = \mL \mS$ can be considered as a form of low-rank matrix approximation from the sparse representation. While the form resembles the low-rank structure used in convex semi-nonnegative matrix factorization~\citep{ding2008convex},  $\mS$ and $\mL$ are not obtained from any matrix factorization algorithm, but are instead direct output from the forward pass of the encoder,  allowing end-to-end training  using SSL objectives.

\subsection{Equivariant Partitioning}

The factorized form in \eqref{eq-lowrank} not only provides a more efficient latent representation ($r(n+d) \ll nd$. With 16 tokens, ViT-base on a $224\times 224$ image enjoys 90\% reduction), it also provides an escape from the invariance paradox. The spatial transformation is now partitioned as follows:

\begin{equation}
    \underbrace{\frac{\partial \mZ(t_\theta \circ X)}{\partial \theta}}_{\text{Total Equivariance}} = 
    \underbrace{\left( \frac{\partial \mL(t_\theta \circ X)}{\partial \theta} \right) \mS}_{\substack{\text{Spatial Equivariance} }} + 
    \underbrace{\mL \left( \frac{\partial \mS(t_\theta \circ X)}{\partial \theta} \right)}_{\text{Semantic Variance} \approx 0}.
\label{eq:factorized_equivariance}
\end{equation}

\begin{figure}
    \centering
    \includegraphics[width=\linewidth]{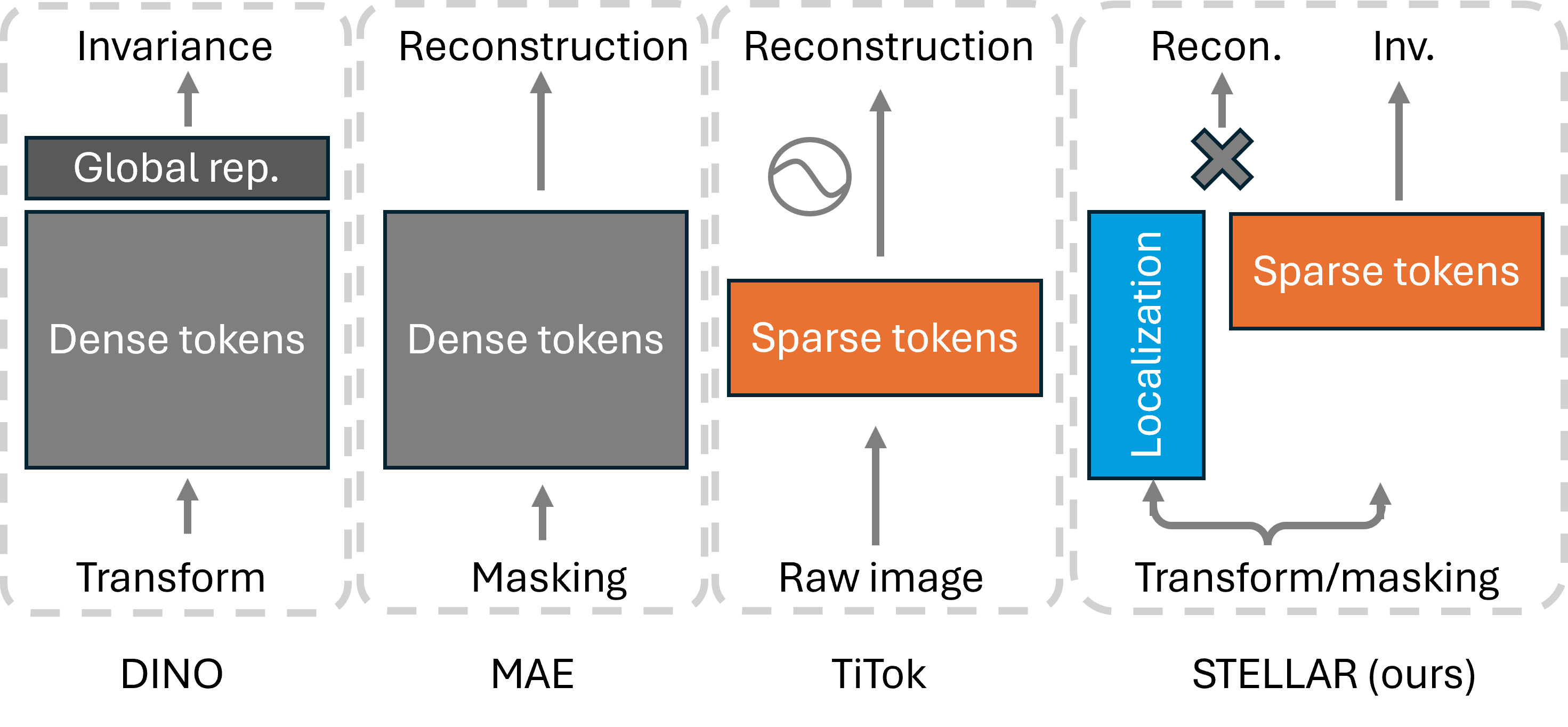}
    \caption{Comparison of learning different latent representation.}
    \label{fig:compare}
\end{figure}

With the spatial and semantic information disentangled, we can offload the spatial equivariance entirely to localization matrix $\mL$, while still achieving semantic invariance in $\mS$. We illustrate the learning paradigm of the factorized representation in Fig.~\ref{fig:compare}. We require that $\mZ(X)$ can reconstruct the image by minimizing 
\begin{equation}
    \mathcal{L}_{recon} = \ell(\mathcal{D}(\mL(X) \mS(X)), X).
\label{eq-recon}
\end{equation}

This \textit{low-rank approximated reconstruction} forces the model to use only sparse tokens $\{\vs_j\}_{j=1}^r$ to capture sufficient information about the image.

% Finally, a compact sparse representation is then obtained by concatenating the concept embeddings with the transposed localization matrix:
% \begin{equation}
%     \mZ = [\mS,\mL^T] \in \mathbb{R}^{r\times d^*}, \quad d^* = d + n.
% \end{equation}

\subsection{Vision Concept Clustering}\label{sec-cluster}
To encourage sparse tokens to represent transferable vision concepts, we structure them into $K$ learnable prototypes $\vc_1, \cdots, \vc_K \in \mathbb{R}^p$. A backbone encoder $\mathcal{E}$ maps a mini-batch of $m$ images into sparse features $\mS^1, \cdots, \mS^m$. Each token is projected onto the unit sphere $\mathbb{S}^{p-1}$ via a normalized projector $h: \mathbb{R}^d \rightarrow \mathbb{S}^{p-1}$, and its similarity to prototypes $\mC=[\vc_1,\cdots,\vc_K]$ gives logits
\begin{equation}\label{eq-logits}
\lambda^i_j = [\vc_1 \cdot h(\vs^i_j), \cdots, \vc_K \cdot h(\vs^i_j)], \quad j=1,\dots,r.
\end{equation}

Soft assignments over the prototypes is obtained with
\begin{equation}\label{eq-assignment}
q^i_{j,k} = \frac{\exp(\lambda^i_{j,k}/\tau)}{\sum_{k'=1}^K \exp(\lambda^i_{j,k'}/\tau)},
\end{equation}
where $\tau$ controls sharpness. Direct entropy minimization of $q^i_j$ is unstable due to non-convexity and empty clusters. Following \cite{caron2020unsupervised, darcet2025cluster}, we compute balanced assignments $\tilde{q}^i_j$ from $q^i_j$ using the Sinkhorn-Knopp algorithm (see appendix) without gradient, and minimize
\begin{equation}\label{eq-cluster}
\mathcal{L}_\text{cluster} = -\frac{1}{mr}\sum_{i=1}^m \sum_{j=1}^r \sum_{k=1}^K \tilde{q}^i_{j,k} \log q^i_{j,k}.
\end{equation}

Unlike DINOv2 and SwAV which only use Sinkhorn for balancing teacher targets, we explicitly minimize $\mathcal{L}_\text{cluster}$ along with all other objectives.

\subsection{Set Concepts Alignment}\label{sec-align}
To achieve the semantic invariance in \eqref{eq:factorized_equivariance}, we align the sparse tokens ${\vs'_1,\dots,\vs'_r}$ obtained from a transformed view (e.g. masking or cropping) to the ones from the global view ${\vs_1,\dots,\vs_r}$. However, this set concepts alignment problem is challenging compared to global representation alignment in traditional JE methods, because there is no inherent ordering in the $r$ tokens. To solve the problem, we apply optimal transport with the cost matrix
\begin{equation}
\Theta_{j'j} = \|\vs'_{j'} - \vs_j\|_2.
\end{equation}
We solve for an assignment matrix $\mP$ via entropy-regularized optimal transport:
\begin{align}\label{eq-match}
\min_{\mP \geq 0} & \quad \sum_{j',j} \mP_{j'j} \Theta_{j'j} - \epsilon H(\mP), \\
s.t. &  \quad
\mP\mathbf{1}_r = \mP^T\mathbf{1}_r = \frac{1}{r}\mathbf{1}_r,
\end{align}
with $H(\mP)=-\sum{j',j} \mP_{j'j}\log \mP_{j'j}$. We solve for $\mP$ using the Sinkhorn algorithm, and define the matching $\sigma(j') := \text{argmax}_j \mP_{j'j}$. Compared to bipartite matching algorithms such Hungarian matching widely used in previous literature, this algorithm is up to $100\times$ faster, with experimental results analyzed in the appendix.

We then compute prototype assignments for the transformed view tokens $q'_{j'} = \text{softmax}(\mC^T h(\vs'_{j'})/\tau)$, and minimize the set concept alignment loss
\begin{equation}\label{eq-align}
\mathcal{L}_\text{align} = -\frac{1}{r}\sum_{j'=1}^r \sum_{k=1}^K \tilde{q}_{\sigma(j'),k}\log q'_{j',k}.
\end{equation}

Optionally, we use the same framework to cluster and align the CLS token with its own projector and prototypes, similar to previous JE methods. However, we do not used it for reconstruction. We also apply KoLeo regularization~\citep{sablayrolles2018spreading} on the normalized sparse tokens $\bar{\vs}_j := \vs_j/\|\vs_j\|$ obtained from the same image to encourage concept diversification: 
\begin{equation}
    \mathcal{L}_{KoLeo} = -\frac{1}{r} \sum_{j=1}^r \log\left(\min_{j' \neq j} \frac{1}{2}\| \bar{\vs}_j - \bar{\vs}_{j'}\|_2 \right).
\end{equation}

All together, we jointly optimize the following objectives by training the encoder $\mathcal{E}$, decoder $\mathcal{D}$, projector $h$, and prototypes $\mC$ jointly with the final objective:
\begin{align}
\min_{\mathcal{E}, \mathcal{D}, h, \mC} \quad & a_1\mathcal{L}_\text{recon} + a_2\mathcal{L}_\text{cluster} + a_3\mathcal{L}_\text{align} + \\ & a_4\mathcal{L}_\text{cluster-cls} + a_5\mathcal{L}_\text{align-cls} + a_6\mathcal{L}_\text{KoLeo}.
\end{align}

In summary, we proposed a sparse vision representation $(\mS, \mL) = \mathcal{E}(X)$ that explicitly disentangles semantic concepts from their spatial distributions, enabling the latent variables to support both pixel-level reconstruction and high-level semantic understanding. We introduced a simple encoder design to obtain these latent variables and SSL objectives to shape them into transferable visual concepts.

We refer to our framework of learning the spatial-semantic factorized representation $\mZ(X) = \mL(X) \mS(X)$ as \textbf{S}parse \textbf{T}oken \textbf{E}xtraction and \textbf{L}ocalization with \textbf{L}ow-rank \textbf{A}pproximated \textbf{R}econstruction (STELLAR). 

\subsection{Model Design}
We note that the framework only specifies the latent space, and does not prescribe any specific encoder or decoder architecture. In this work, we adopt a simple design with common modules and model architectures to obtain $\mS$ and $\mL$ as described below.

For the encoder part, we use an existing ViT\citep{dosovitskiy2020image} as the backbone, and equip it with $r$ learnable latent query vectors, which are passed to the transformer blocks alongside the patch tokens. Processed by the ViT jointly, the latent queries produce sparse tokens $\mS \in \mathbb{R}^{r \times d}$. 

% In order to} obtain the localization matrix $\mL \in \mathbb{R}^{n \times r}$ associated with the sparse tokens, we simply project $\mS$ and $\mU$ to a shared space, and} compute pairwise cosine similarities between projected dense and sparse features, followed by a softmax normalization with temperature $t$:
% \begin{equation}
%     \mL = \text{softmax}\left(1/t \cdot \text{cossim}(\mU \mW_1, \mS \mW_2)\right),
% \end{equation}
% where $\mW_1$ and $\mW_2$ are learnable linear projections, and $t$ controls the sharpness of the spatial distribution. We note that this form is similar to the attention weights in a single head attention from the dense tokens to the sparse tokens, despite L2 normalization and dedicated softmax temperature. This simple form was proven to be effective in obtaining $\mL$ in our experiments.}

To obtain the localization matrix $\mL \in \mathbb{R}^{n \times r}$ associated with the sparse tokens, we use the dense feature map $\mU \in \mathbb{R}^{n \times d}$ output from the image patches. We project both $\mS$ and $\mU$ into a shared embedding space and compute their pairwise cosine similarities, followed by a softmax normalization with temperature $\tau_\text{spatial}$ along the the second dimension:
\begin{equation}
    \mL = \text{softmax}\left(\text{cossim}(\mU \mW_1, \mS \mW_2)/\tau_\text{spatial} \right).
\end{equation}

\begin{figure*}[!hbt]
    \centering
    
    \includegraphics[width=\textwidth]{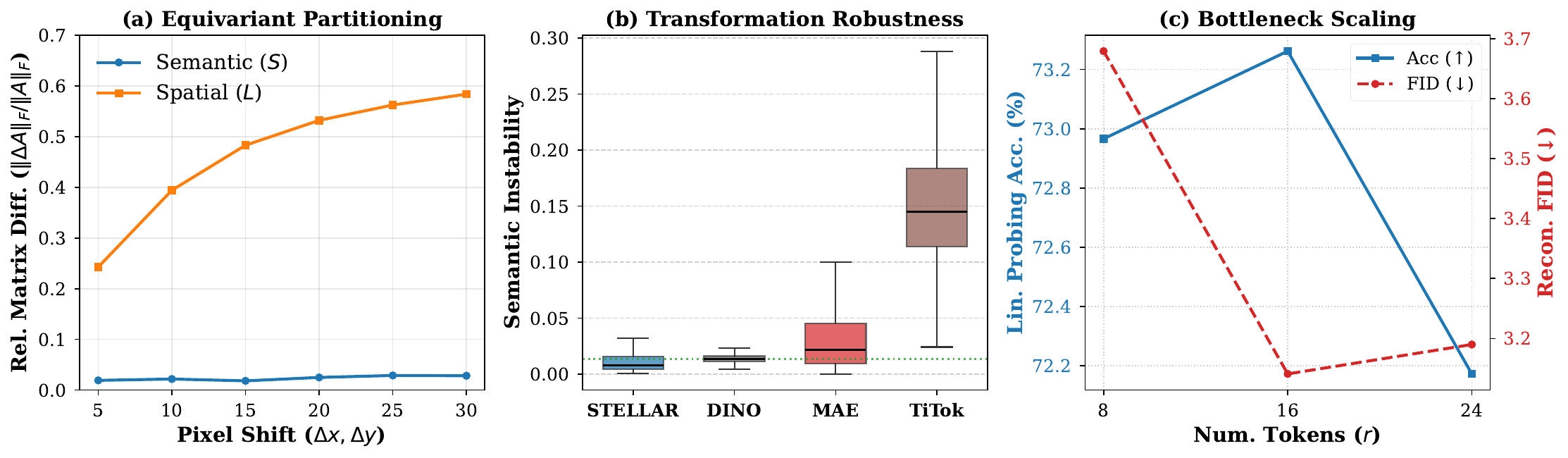}
    \caption{Analysis of STELLAR representation. (a) Relative matrix difference in $\mL$ and $\mS$ under controled pixel shift in the input image. (b) Cosine distance of latent representation under random 50-100\% random cropping. (c) Impact of number of sparse tokens $r$ on reconstruciton and semantic quality.}
    \label{fig:analysis}
\end{figure*}

$\mW_1$ and $\mW_2$ are learnable linear projections, and $\tau_\text{spatial}$ controls the sharpness of the spatial distribution. We note that his mapping is structurally similar to the attention weights obtained in a single-head cross-attention layer, up to the use of L2 normalization and an explicit temperature parameter. Therefore, the latent representation  $\mZ = \mL \mS$ can be viewed as rebuilding a dense feature map for reconstruction by cross-attending to only $r$ sparse concept tokens.

All together, the encoder $\mathcal{E}$ includes ViT transformer blocks, $r$ learnable latent query vectors, and projection layers $\mW_1, \mW_2$. The decoder $\mathcal{D}$ is a 6-layer lightweight ViT reconstructing the image patches.

The STELLAR framework can be used on a pretrained ViT such as MAE or DINO to leverage the foundation prior and shape it into a sparse holistic representation. It can also be trained from a random prior and reach competitive spatial, semantic, and reconstruction quality. We provide deep analysis in the ablation study.

% Although the low-rank form $\mL \mS$ is simple, it effectively disentangles high-level semantic concepts from low-level spatial localization. This yields two key benefits for both reconstruction and representation learning:

% \begin{itemize}
% \item The concept matrix $\mS$ no longer needs to encode spatial information, and can instead focus purely on learning what objects or visual concepts are present. Through the linear combination $\mL \mS$, these concepts can be flexibly allocated across spatial locations to form a dense semantic map, enabling efficient reconstruction.

%     \item Because the semantic embeddings in $\mS$ are independent of location, we can freely apply image transformations while enforcing consistency in the learned concepts. This invariance induces robust high-level semantic features that transfer well to image understanding tasks.  

% \end{itemize}

% We cluster the sparse visual concepts from the dataset into prototypes in an online manner. Random partial views (masking, cropping) are passed to the same encoder, and aligned to the global view using set optimal transport.

% In addition, we optionally use an exponential moving average (EMA) updated momentum encoder to encode the target assignments $\tilde{q}$ in \eqref{eq-cluster} and \eqref{eq-align}. We observed that using a momentum encoder is essential in the warm-up stage when training from scratch, but suboptimal in subsequent training. We provide detailed results in ablation study. 
\section{Experiments}

We train STELLAR on ImageNet-1K~\citep{deng2009imagenet} without labels. The encoder is a vanilla ViT~\citep{dosovitskiy2020image} augmented with $8$–$24$ learnable latent queries that produce sparse tokens. A lightweight $6$-layer ViT serves as the decoder predicting either MaskGIT-VQGAN tokens~\citep{esser2021taming, chang2022maskgit}. When using a foundation prior for the backbone ViT, we ensure that the pretraining was also performed only on ImageNet-1K. We use MAE as the default prior and studied the effect of different prior modes in ablation study. When training from random prior, we use a momentum updated encoder to encode the target assignments $\tilde{q}$ in \eqref{eq-cluster} and \eqref{eq-align}, following \citeauthor{grill2020bootstrap, caron2021emerging}. 

\begin{figure*}
    \centering
    \includegraphics[width=0.49\linewidth]{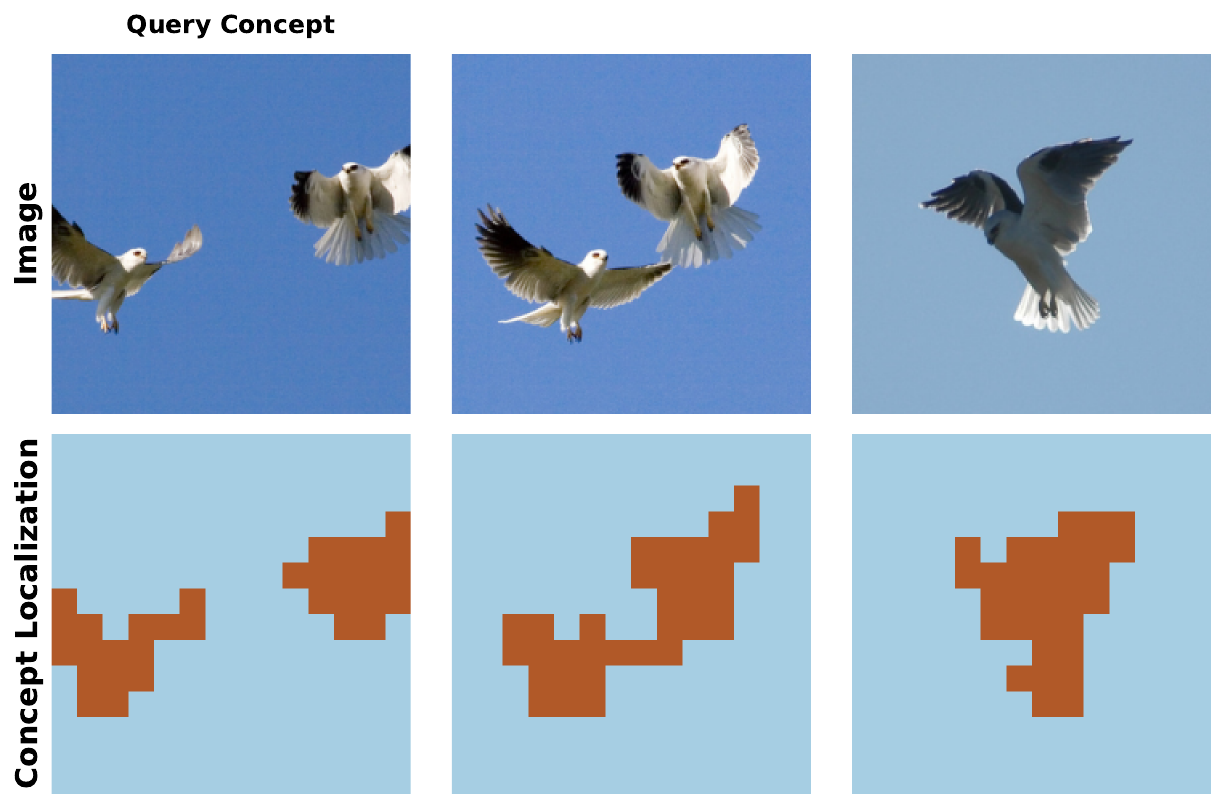}
    \includegraphics[width=0.49\linewidth]{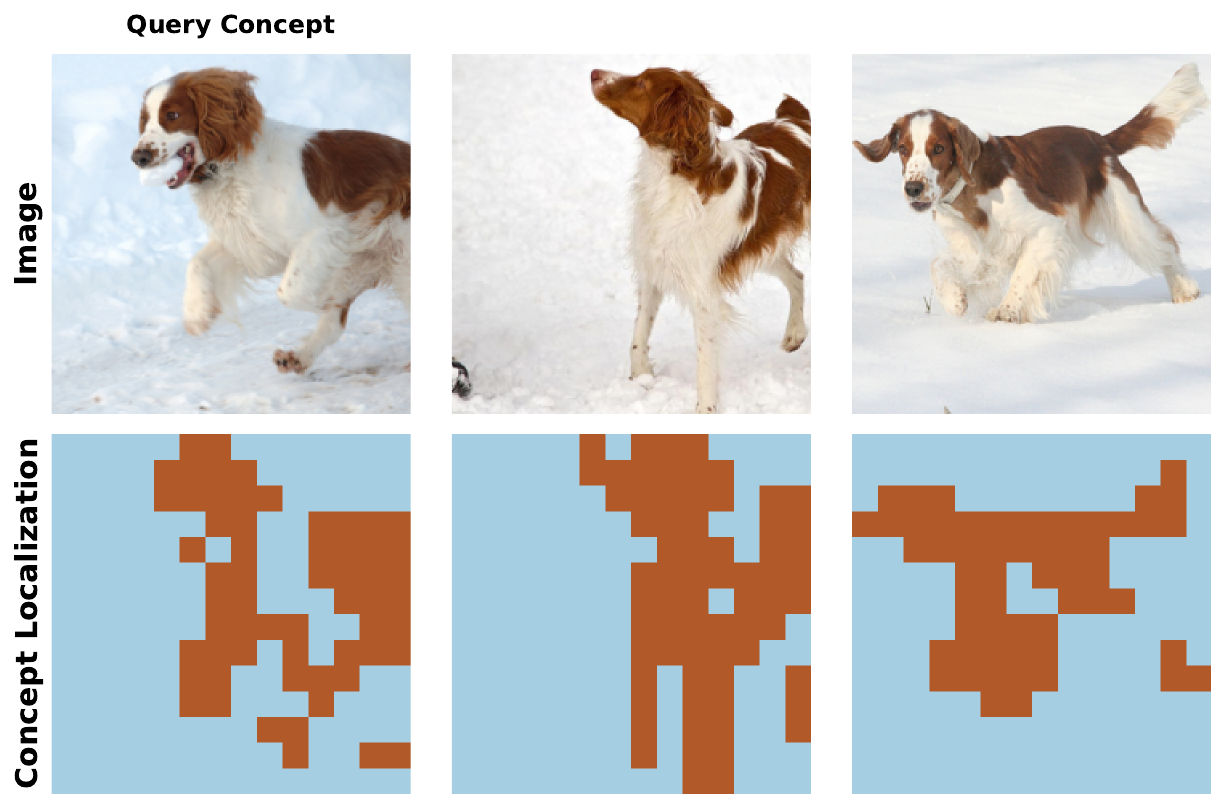}
    \caption{Vision concepts retrieved from the training set. We show show both the image and the spatial localization of the concept.}
    \label{fig:retrieve}
\end{figure*}

\subsection{Probing the Factorized Representation}

We designed a series of experiments to analyzed the factorized representation $\mZ = \mL \mS$ from STELLAR training. 

\paragraph{Experiment 1: Equivariant Partitioning}
We built a controllable and parametrized spatial transformation group to examine the equivariant partitioning in \eqref{eq:factorized_equivariance}. Given an image, we take a crop and shift it gradually from 5 to 30 pixel, either horizontally or vertically. We calculated the relative matrix difference $\frac{\|\mS(t_\theta \circ X) - \mS( X)\|_F}{\|\mS( X)\|_F}$ and $\frac{\|\mL(t_\theta \circ X) - \mL( X)\|_F}{\|\mL( X)\|_F}$. Optimal tsansport in \eqref{eq-match} is used to match the token ordering. As shown in Fig.~\ref{fig:analysis}(a), the semantic matrix $\mS$ stays almost completely invariant, while the spatial localization matrix $\mL$ changes continuously with the spatial shift, proving the effectiveness of equivariant partitioning in the factorized representation.
\paragraph{Experiment 2: Transformation Robustness} 
Next we compare the semantic stability of STELLAR representation with baseline models under random resized cropping at scale 50-100\%. We calculate the cosine distance from the feature of the untransformed image as a measure of semantic instability. For DINO and MAE, we used mean-pooled dense features, and for sparse models STELLAR and TiTok, we used mean-pooled sparse tokens. As shown in Fig.~\ref{fig:analysis}(b), the sparse tokens of STELLAR enjoys high transformation robustness at DINO level. As expected, the reconstruction-based models MAE and TiTok show higher variance to spatial transformation. Specifically, the un-factorized sparse representation from TiTok is extremely unstable, as the model need to store both semantic and spatial information in the same sparse tokens for reconstruction.
\paragraph{Experiment 3: Effect of Low-rank Bottleneck} 
The number of sparse tokens $r$ serves as the intrinsic rank of the latent representation. We experimented scaling $r$
from 8 to 24, and evaluated reconstruction with FID~\citep{NIPS2017_8a1d6947} and semantics with linear probing on mean-pooled sparse tokens. As shown in Fig.~\ref{fig:analysis}(c), the linear probing accuracy decreases as $r$ increases, while reconstruction improves with more tokens, showing a trade-off in the intrinsic rank of the representation. As a sweet spot, $r=16$ enjoys both rich semantics and high-quality reconstruction, which we used as the default for all other experiments.

Finally, we visualize the factorized representation in Fig.~\ref{fig:retrieve}. We show the spatial localization (thresholded by $1/r$) of a sparse token in the image, and the top retrieved semantic concepts from the training dataset.

\subsection{Evaluating Holistic Representation}

Next we examine the semantic quality and reconstruction potential of representation from different models. We trained the same decoder in STELLAR on top of the frozen features from different encoders. The original decoder in STELLAR was also finetuned with the rest of the model frozen. TiTok used it's native decoder, which is a full-sized ViT compared to only 6 layers in STELLAR.

As shown in Table~\ref{fid-acc-table}, STELLAR (shown as ours) shows superior performance in supporting both semantics and reconstruction. The linear probing and k-NN accuracy of STELLAR surpass reconstruction-feasible representations from all baselines, despite trailing behind the CLS token from DINO, which is infeasible as a reconstruction latent.

% Table \ref{fid-acc-table} and Figure \ref{fig:tokens} shows results across token counts ($8$, $16$, $24$), compared with TiTok~\citep{yu2024image} and MAE~\citep{he2022masked}. STELLAR achieves strong reconstruction even with few tokens (rFID $=3.68$ with $8$ tokens; $2.60$ with ViT-H, $16$ tokens), approaching MaskGIT-VQGAN ($2.28$) without decoder finetuning. For linear probing, STELLAR maintains robust accuracy and does not drop with more tokens as in TiTok~\citep{yu2024image}. Reconstruction improves with the number of token, but plateau after 16. We used 16 tokens in all other experiments unless specified. We also see in Figure \ref{fig:tokens} that STELLAR preserves the location of the objects. Interestingly, it also automatically removes the dark edge in the bottom example, indicating it is reconstructing from high-level semantics rather than memorizing low-level details. Overall, STELLAR balances efficient reconstruction and discriminative understanding at high quality.

\begin{table}[h]
\caption{Reconstruction and semantic metrics on IN1K of STELLAR (ours) and baseline models. For reference, we also reported semantic metrics of the global representation from DINO, and huge size STELLAR model. Best main results are shown in \textbf{bold}. Model sizes are ViT-B by default, with larger sizes indicated in parentheses. *: TiTok used its native ViT decoder of larger size.}
\label{fid-acc-table}
\begin{center}
\begin{small}
\begin{sc}
\begin{tabular}{lc|cc|cc}
\toprule
\multicolumn{2}{c}{} & \multicolumn{2}{c}{Reconstruction} & \multicolumn{2}{c}{Semantics} \\
\midrule
Model  & \# tks & FID $\downarrow$ & LPIPS $\downarrow$ & Lin. & kNN \\
\midrule
DINO & 1 & - & - & \color{gray}{76.46}& \color{gray}{74.69}\\
DINO & 196 & 3.27 & 0.2121 & 70.31 & 54.41 \\
MAE & 196 & 3.02 & \textbf{0.2071}& 66.32 & 25.82 \\
TiTok* & 32 & 2.75 & 0.3281 & 33.42 & 7.30 \\
TiTok* & 64 & 1.99 & 0.2571 & 32.87 & 7.29 \\
\midrule
ours & 16 & 3.06 & 0.2077 & \textbf{73.26}& \textbf{67.25}\\
ours & 196 & \textbf{2.85}& 0.2085 & 72.21 & 64.71 \\
ours(H)& 16 & 2.60& 0.1729& 79.10& 77.31\\
\bottomrule
\end{tabular}
\end{sc}
\end{small}
\end{center}
\vskip -0.1in
\end{table}

On reconstruction, STELLAR shows comparable FID and LPIPS loss~\cite{zhang2018unreasonable} to the dense feature map from MAE, with 90\% reduction in latent size. Although TiTok achieved lower FID with a much larger decoder, its shows highest LPIPS loss, indicating poor spatial consistency. In contrast, STELLAR exhibits superior reconstruction locality even with fewer tokens. The full-rank dense feature map $\mU$ (196 tokens) from the ViT in STELLAR shows even lower reconstruction FID, while drops in semantic quality. Finally, when scaling to huge-sized ViT, STELLAR achieves top reconstruction and semantic quality, even without decoder finetuning.

\subsection{Benchmarking Image Understanding}

Lastly, we benchmark STELLAR in classical image understanding tasks with linear probing on frozen features, comparing against other ImageNet-pretrained SSL models. We report results for classification on ImageNet-1K (IN1K), Oxford-IIIT Pet (Pets)~\citep{parkhi2012cats}, Food-101 (Food)~\citep{bossard14}, and GlaS~\citep{sirinukunwattana2016gland} for cancer grade classification in histopathology. Segmentation benchmarks include ADE20K~\citep{zhou2017scene}, Cityscapes~\citep{cordts2016cityscapes}, and Pascal VOC~\citep{everingham2010pascal}.  For broader context, we also include AIM~\citep{el2024scalable} and DINOv2~\citep{oquab2023dinov2}, which leverage substantially larger training corpora (100–1000$\times$ more images).

\begin{table*}[t]
\caption{\textbf{Evaluation of Fine-grained and Global Image Understanding.} We evaluate semantic segmentation (mIoU \%) and classification accuracy (\%) via linear probing on frozen features. We used the dense feature map from the backbone for all segmentation tasks and all models. \textbf{Bold}: best with ImageNet training. \underline{Underline}: best in architectural class (e.g., ViT-B).}
\label{tab:merged_results}
\begin{center}
\begin{small}
\begin{sc}
\begin{tabular}{l l c c | c c c | c c c c}
\toprule
& & \multicolumn{2}{c|}{\bf SSL Type} & \multicolumn{3}{c|}{\bf Segmentation (mIoU)} & \multicolumn{4}{c}{\bf Classification (Acc)} \\
\bf Model & \bf Arch. & \bf Target & \bf Method & \bf ADE20K & \bf CitySc & \bf VOC & \bf IN1K & \bf Pets & \bf Food & \bf GlaS \\
\midrule
\multicolumn{11}{l}{\it Semantic-Centric (Joint Embedding / Invariance)} \\
BYOL & RN-50 & Global & Distill & 18.43 & \underline{18.66} & 63.89 & \underline{70.39} & \underline{82.77} & \underline{64.57} & \underline{95.00} \\
MoCo v3 & ViT-B & Global & Contr. & 29.45 & 25.13 & 74.08 & 74.31 & 91.14 & 77.47 & \textbf{97.50} \\
DINO & ViT-B & Global & Distill & 26.87 & 26.82 & 79.29 & \underline{76.46} & \textbf{93.84} & \underline{79.28} & 95.00 \\
MSN & ViT-B & Global & Masking & 26.66 & 25.39 & 68.59 & 73.65 & 75.91 & 68.93 & 92.50 \\
DenseCL & RN-50 & Dense & Contr. & \underline{23.08} & 18.63 & \underline{70.95} & 61.10 & 72.99 & 59.16 & 85.00 \\
Data2Vec & ViT-B & Dense & Lat-MIM & 22.03 & 23.49 & 61.33 & 54.90 & 26.47 & 34.40 & 73.75 \\
SiameseIM & ViT-B & Dense & Lat-MIM & 29.24 & 26.52 & 81.38 & 74.97 & 91.61 & 71.01 & 91.25 \\
I-JEPA & ViT-H & Dense & Lat-MIM & 21.57 & 18.59 & 74.13 & 71.72 & 84.68 & 70.34 & 87.50 \\
iBOT & ViT-B & Gl+De & Dist+MIM & \underline{31.78} & 25.69 & 77.06 & 76.40 & 92.40 & 78.08 & 96.25 \\
iBOT & ViT-L & Gl+De & Dist+MIM & 33.26 & 26.37 & 77.57 & \underline{78.53} & 92.12 & \textbf{81.07} & 96.25\\
\midrule
\multicolumn{11}{l}{\it Image-Centric (Reconstruction)} \\
BEIT & ViT-B & Dense & Tok MIM & 11.58 & 18.90 & 27.44 & 32.94 & 36.20 & 54.49 & 90.00 \\
BEIT & ViT-L & Dense & Tok MIM & 12.64 & 20.37 & 25.48 & 36.77 & 36.71 & 56.03 & 90.00 \\
SimMIM & Swin-B & Dense & Pix MIM & 12.46 & 17.23 & 35.14 & 24.77 & 27.39 & 40.94 & 77.50 \\
MAE & ViT-B & Dense & Pix MIM & 30.91 & \underline{29.44} & 76.43 & 66.32 & 81.58 & 70.40 & 93.75 \\
MAE & ViT-L & Dense & Pix MIM & \underline{34.36} & \underline{32.53} & 77.79 & 73.09 & 84.30 & 76.22 & 95.00 \\
MAE & ViT-H & Dense & Pix MIM & 36.16 & \bf{35.21} & 78.07 & 75.22 & 84.96 & \underline{78.36} & \underline{95.00} \\
SemMAE & ViT-B & Dense & Pix MIM & 3.52 & 25.48 & 48.33 & 43.84 & 56.99 & 58.90 & 92.50 \\
TiTok-64 & ViT-B & Sparse & Sprs Rec & -- & -- & -- & 32.87 & 42.06 & 43.68 & \textbf{97.50} \\
TiTok-32 & ViT-L & Sparse & Sprs Rec & -- & -- & -- & 33.42 & 27.83 & 38.83 & 78.75 \\
\midrule
\multicolumn{11}{l}{\it Our Method (Sparse Factorized Modeling)} \\
\bf STELLAR & ViT-B & Sparse & Inv+Rec & 31.33 & 27.74 & \underline{81.83} & 73.26 & 89.70 & 74.09 & 95.00 \\
\bf STELLAR & ViT-L & Sparse & Inv+Rec & 34.02 & 31.32 & \textbf{85.90} & 76.94 & \underline{92.53} & 74.78 & \textbf{97.50} \\
\bf STELLAR & ViT-H & Sparse & Inv+Rec & \textbf{36.66} & 33.30 & \underline{85.66} & \textbf{79.10} & \underline{92.53} & 77.43 & 92.50 \\
\midrule
\multicolumn{11}{l}{\it Larger Scale Pretraining Beyond ImageNet (Reference Only)} \\
AIM & 600 M & Dense & Image AR & 29.00 & 27.04 & 64.55 & 63.78 & 64.68 & 75.19 & 98.75 \\
AIM & 1 B & Dense & Image AR & 29.59 & 27.05 & 63.90 & 66.86 & 64.21 & 77.96 & 96.25 \\
DINOv2 & ViT-B* & Gl+De & Dist+MIM & 40.10 & 34.66 & 89.52 & 82.82 & 95.59 & 91.08 & 98.75 \\
DINOv2 & ViT-L* & Gl+De & Dist+MIM & 40.45 & 32.07 & 89.19 & 84.23 & 96.08 & 92.94 & 98.75 \\
\bottomrule
\end{tabular}
\end{sc}
\end{small}
\end{center}
\vskip -0.1in
\end{table*}

As shown in Table \ref{tab:merged_results}, the feature map from STELLAR achieves superior performance on ADE20K and Pascal VOC, showing strong fine-grained understanding despite not applying SSL objectives directly to the dense feature map $\mU$. Sparse token modeling implicitly organizes the feature map into semantic regions: to reconstruct the image, each token must encode information covering all spatial parts of the scene, resulting in region-aware representations. While MAE leads in CityScapes, STELLAR follows closed with performance comparable to MAE and DINOv2.

On global image understanding tasks, STELLAR achieves the highest accuracy on IN1K at large model scale, but smaller variants underperform methods such as DINO, which explicitly optimizes for global representations. In general, STELLAR outperforms image reconstruction models and most JE methods, but trails behind top JE models in global semantics. As we do not model the image as a single concept, averaging token features can dilute discriminative information, which is particularly detrimental on object-centric datasets like Pets and Food. Interestingly, on histopathology images involving complex tissue microenvironments, STELLAR achieves the best performance. These results indicate that STELLAR excels at modeling complex, multi-object scenes, while global classification on simple object-centric datasets remains more challenging.

% \begin{table*}[t]
% \caption{\small Ablation on IN1K linear probing accuracy (\%) and ADE20K linear probing mIOU (\%).}
% \label{ablation-table}
% \begin{center}
% \begin{small}
% \begin{sc}
% \begin{tabular}{ccccc|ccc}
% \multicolumn{1}{c}{Recon.}  &\multicolumn{1}{c}{Cluster} &  Set align & CLS align & KoLeo & Recon FID $\downarrow$  & IN1K lin. & ADE20K lin. \\
% \hline 
% \multicolumn{7}{l}{\it Default} \\
%    \checkmark    & \checkmark  & \checkmark & \checkmark & \checkmark & 3.14 & 73.26 & 31.33\\
% \hline 
% \multicolumn{7}{l}{\it Ablation model versions} \\
%  \ding{55}   & \checkmark  & \checkmark & \checkmark & \checkmark & - & 72.44 (-0.82) & 29.94 (-1.39) \\
%  \checkmark    &  \ding{55}   &  \ding{55}  &  \ding{55}  & \checkmark & 3.21 (+0.07) & 52.07 (-21.19) & 20.46 (-10.87) \\
%  \checkmark    & \checkmark  & \ding{55}  & \ding{55}  & \checkmark & 8.95 (+5.81) & 2.73 (-70.53) & 1.93 (-29.39) \\
%  \checkmark    &  \ding{55} & \checkmark  & \checkmark  & \checkmark & 3.62 (+0.48) & 42.14 (-31.12) & 18.90 (-12.43) \\
%   \checkmark    & \checkmark  & \checkmark & \ding{55}  & \checkmark & 3.26 (+0.12) & 70.79 (-2.47) & 30.20 (-1.12) \\
%   \checkmark    & \checkmark  & \checkmark & \checkmark & \ding{55}  & 3.25 (+0.11) & 72.05 (-1.21) & 30.10 (-1.23) \\
% \end{tabular}
% \end{sc}
% \end{small}
% \end{center}
% \vskip -0.1in
% \end{table*}

\begin{table*}[t]
\caption{\textbf{Ablation.} We isolate the impact of each objective on semantic abstraction (IN1K) and spatial grounding (ADE20K), and reconstruction (FID). \textit{Default} denotes the full STELLAR framework. All results are based on ViT-B.}
\label{ablation-table}
\begin{center}
\begin{small}
\begin{sc}
\begin{tabular}{l ccccc | ccc}
\toprule
& \bf Recon. & \bf Cluster & \bf Set Align& \bf CLS Align& \bf KoLeo & \bf rFID $\downarrow$ & \bf IN1K $\uparrow$ & \bf ADE $\uparrow$ \\
\midrule
Default & \checkmark & \checkmark & \checkmark & \checkmark & \checkmark & \bf 3.14 & \bf 73.26 & \bf 31.33 \\
\midrule
\multicolumn{9}{l}{\it Impact of Individual Components} \\
(a) & \ding{55} & \checkmark & \checkmark & \checkmark & \checkmark & --- & 72.44 \color{gray}{(-0.82)} & 29.94 \color{gray}{(-1.39)} \\
(b) & \checkmark & \ding{55} & \ding{55} & \ding{55} & \checkmark & 3.21 \color{gray}{(+0.07)} & 52.07 \color{red}{(-21.19)} & 20.46 \color{red}{(-10.87)} \\
(c) & \checkmark & \checkmark & \ding{55} & \ding{55} & \checkmark & 8.95 \color{red}{(+5.81)} & 2.73 \color{red}{(-70.53)} & 1.93 \color{red}{(-29.39)} \\
(d) & \checkmark & \ding{55} & \checkmark & \checkmark & \checkmark & 3.62 \color{gray}{(+0.48)} & 42.14 \color{red}{(-31.12)} & 18.90 \color{red}{(-12.43)} \\
(e) & \checkmark & \checkmark & \checkmark & \ding{55} & \checkmark & 3.26 \color{gray}{(+0.12)} & 70.79 \color{gray}{(-2.47)} & 30.20 \color{gray}{(-1.12)} \\
(f) & \checkmark & \checkmark & \checkmark & \checkmark & \ding{55} & 3.25 \color{gray}{(+0.11)} & 72.05 \color{gray}{(-1.21)} & 30.10 \color{gray}{(-1.23)} \\
\bottomrule
\end{tabular}
\end{sc}
\end{small}
\end{center}
\vskip -0.1in
\end{table*}

\subsection{Ablation Analysis}
\textbf{Low-rank approximated reconstruction.}
As shown in Table~\ref{ablation-table}, removing the low-rank reconstruction objective (A) reduces both global and fine-grained understanding. Since the remaining objectives resemble typical SSL methods, the model still retains reasonable global performance, but fine-grained understanding suffers more. This indicates that low-rank reconstruction encourages sparse tokens to serve as holistic representations covering the entire image.

\textbf{Concept clustering.}
Eliminating online clustering and set alignment (B) leads to a sharp drop in understanding, highlighting the necessity of structuring sparse tokens into view-invariant concepts. Even when the alignment loss is present (D), missing the clustering loss still lead to feature collapse. 

\textbf{Set alignment.}
The training collapsed when training with only reconstruction and clustering (C), underscoring the critical role of set concepts alignment. Additional alignment on the CLS token (E) primarily benefits global classification but has limited effect on spatial grounding. Finally, KoLeo regularization (F) consistently improves all tasks at similar level. Interestingly, the absent of either concept clustering or set alignment led to a sharp drop in performance.

\textbf{Foundational Prior} We ablated STELLAR trained from different pretrained foundational prior in Table \ref{init-table}. We observes that STELLAR significantly boost the semantic quality from MAE, and the spatial grounding from DINO prior. The semantics performance falls to similar level despite different foundational priors. When training from random prior, STELLAR is able to reach semantics at MAE level and spatial understanding similar to that from DINO prior. The reconstruction quality stays consistent in all cases.

% \begin{table*}[ht]
% \caption{Ablation on model initialization and training strategy}
% \label{init-table}
% \begin{center}
% \begin{small}
% \begin{sc}
% \begin{tabular}{c|cc|ccccccccc}
% \multicolumn{1}{c}{} & \color{gray}{MAE} & \color{gray}{DINO}  & MAE init. & DINO init. & EMA only & EMA+75ep. \\
% \hline 
% Recon. FID ($\downarrow$) & - & - & 3.14 & 3.31 & 3.69 & 3.21 \\
% Class. Acc. & \color{gray}{66.32} & \color{gray}{76.46} &  73.26 & 73.31 & 58.83  & 65.28 \\
% Seg. mIOU & \color{gray}{30.91} & \color{gray}{26.87} &  31.33 & 28.17 & 26.79 & 28.10 \\
% \end{tabular}
% \end{sc}
% \end{small}
% \end{center}
% \end{table*}

\begin{table}[ht]
\caption{Evaluating STELLAR trained from different foundational priors. \textit{Base} represents the performance of the original backbone.}
\label{init-table}
\begin{center}
\begin{small}
\begin{sc}
\setlength{\tabcolsep}{4pt}
\begin{tabular}{l | c | cc | cc}
\toprule
&  \bf Recon & \multicolumn{2}{c|}{\bf Semantic (IN1K)} & \multicolumn{2}{c}{\bf Spatial (ADE20K)} \\
\bf Prior &  \bf FID $\downarrow$ & Base & \bf +STELLAR & Base & \bf +STELLAR \\
\midrule
MAE &  \bf 3.14 & 66.32 & \bf 73.26 \color{blue}{(+6.9)} & 30.91 & \bf 31.33 \color{blue}{(+0.4)} \\
DINO &  3.31 & 76.46 & 73.31 \color{red}{(-3.2)} & 26.87 & \bf 28.17 \color{blue}{(+1.3)} \\
\midrule
rand & 3.21 & -- & 65.28 & -- & 28.10 \\
\bottomrule
\end{tabular}
\end{sc}
\end{small}
\end{center}
\end{table}
\section{Discussion and Concluding Remarks}

We have demonstrated that the long-standing trade-off between \textit{semantic abstraction} and \textit{spatial grounding} (the Invariance Paradox) is largely an artifact of the traditional dense-grid representation. By factorizing the latent space into sparse ``What'' and ``Where'' components, STELLAR effectively resolves this conflict.

The core of STELLAR’s success lies in the principle of \textit{semantic triage}. In a dense model (e.g., MAE), the representation is spatially exhaustive but semantically diluted; the model is forced to allocate representational capacity to every patch, including stochastically redundant background noise.

The low-rank factorization provides the mathematical machinery to disentangle two distinct types of visual information. The \textit{Concept Tokens} ($\mathbf{S}$) are trained to be view-invariant, capturing the categorical ``What,'' while the \textit{Spatial Coefficients} ($\mathbf{L}$) remain equivariant, capturing the geometric ``Where.'' 

This disentanglement allows the model to satisfy the Joint Embedding objective (alignment across views) without destroying the spatial anchors needed for high-fidelity reconstruction. By separating these factors, we avoid the ``semantic blurring'' often seen in global-pooling methods, as each sparse token maintains a precise, albeit flexible, relationship with the physical image geometry.

\textbf{The Path to Unified Multimodality}: Perhaps the most promising frontier for STELLAR is its potential as a visual front-end for Large Language Models (LLMs). Because our tokens are sparse and semantically grounded, they offer a more natural interface for cross-modal alignment than the hundreds of dense tokens generated by standard ViTs. Future work will investigate the systematic integration of STELLAR concepts with linguistic embeddings for more interpretable multimodal understanding.

Overall, our results highlight sparse tokens as a promising direction for unifying efficiency, interpretability, and semantic richness in self-supervised representation learning.

% \section{Discussion and Conclusion}

% We demonstrate that sparse visual representation, when structured in form of low-rank factorization, can simultaneously support high-fidelity reconstruction and rich semantic understanding. By disentangling \textit{what} and \textit{where}, STELLAR learns compact tokens that transfer effectively to both global and fine-grained tasks, surpassing prior methods. The sparse image modeling framework effectively shapes the feature map into semantic regions, without direct loss on the dense representation.

% Nevertheless, several limitations remain. First, our experiments are conducted primarily on ImageNet-1K scale; extending to larger pretraining corpora is likely to further narrowing the gap with large-scale models such as DINOv2. Second, the current framework adopts a minimalist architecture. Exploring more expressive decoders or hybrid architectures could enhance reconstruction fidelity. Finally, while STELLAR naturally provides a compact interface for multimodal alignment, we only experimented simple alignment in appendix, leaving systematic integration with language models to future work. Overall, our results highlight sparse tokens as a promising direction for unifying efficiency, interpretability, and semantic richness in representation learning.

\section*{Impact Statement}

This paper presents work whose goal is to advance the field of machine learning. There are many potential societal consequences of our work, none of which we feel must be specifically highlighted here.

% In the unusual situation where you want a paper to appear in the
% references without citing it in the main text, use \nocite
% \nocite{langley00}

\bibliography{main}

@ARTICLE{CNN,
  author={Lecun, Y. and Bottou, L. and Bengio, Y. and Haffner, P.},
  journal={Proceedings of the IEEE}, 
  title={Gradient-based learning applied to document recognition}, 
  year={1998},
  volume={86},
  number={11},
  pages={2278-2324},
  doi={10.1109/5.726791}}

@article{bengio2013representation,
  title={Representation learning: A review and new perspectives},
  author={Bengio, Yoshua and Courville, Aaron and Vincent, Pascal},
  journal={IEEE transactions on pattern analysis and machine intelligence},
  volume={35},
  number={8},
  pages={1798--1828},
  year={2013},
  publisher={IEEE}
}

@inproceedings{sun2021sparse,
  title={Sparse r-cnn: End-to-end object detection with learnable proposals},
  author={Sun, Peize and Zhang, Rufeng and Jiang, Yi and Kong, Tao and Xu, Chenfeng and Zhan, Wei and Tomizuka, Masayoshi and Li, Lei and Yuan, Zehuan and Wang, Changhu and others},
  booktitle={Proceedings of the IEEE/CVF conference on computer vision and pattern recognition},
  pages={14454--14463},
  year={2021}
}

@inproceedings{he2016deep,
  title={Deep residual learning for image recognition},
  author={He, Kaiming and Zhang, Xiangyu and Ren, Shaoqing and Sun, Jian},
  booktitle={Proceedings of the IEEE conference on computer vision and pattern recognition},
  pages={770--778},
  year={2016}
}

@inproceedings{he2020momentum,
  title={Momentum contrast for unsupervised visual representation learning},
  author={He, Kaiming and Fan, Haoqi and Wu, Yuxin and Xie, Saining and Girshick, Ross},
  booktitle={Proceedings of the IEEE/CVF Conference on Computer Vision and Pattern Recognition},
  pages={9729--9738},
  year={2020}
}

@inproceedings{wang2021dense,
  title={Dense contrastive learning for self-supervised visual pre-training},
  author={Wang, Xinlong and Zhang, Rufeng and Shen, Chunhua and Kong, Tao and Li, Lei},
  booktitle={Proceedings of the IEEE/CVF Conference on Computer Vision and Pattern Recognition},
  pages={3024--3033},
  year={2021}
}

@article{dosovitskiy2020image,
  title={An image is worth 16x16 words: Transformers for image recognition at scale},
  author={Dosovitskiy, Alexey and Beyer, Lucas and Kolesnikov, Alexander and Weissenborn, Dirk and Zhai, Xiaohua and Unterthiner, Thomas and Dehghani, Mostafa and Minderer, Matthias and Heigold, Georg and Gelly, Sylvain and others},
  journal={arXiv preprint arXiv:2010.11929},
  year={2020}
}

@article{bao2021beit,
  title={Beit: {BERT} pre-training of image transformers},
  author={Bao, Hangbo and Dong, Li and Wei, Furu},
  journal={arXiv preprint arXiv:2106.08254},
  year={2021}
}

@inproceedings{he2022masked,
  title={Masked autoencoders are scalable vision learners},
  author={He, Kaiming and Chen, Xinlei and Xie, Saining and Li, Yanghao and Doll{\'a}r, Piotr and Girshick, Ross},
  booktitle={Proceedings of the IEEE/CVF Conference on Computer Vision and Pattern Recognition},
  pages={16000--16009},
  year={2022}
}

@inproceedings{cheng2022masked,
  title={Masked-attention mask transformer for universal image segmentation},
  author={Cheng, Bowen and Misra, Ishan and Schwing, Alexander G and Kirillov, Alexander and Girdhar, Rohit},
  booktitle={Proceedings of the IEEE/CVF conference on computer vision and pattern recognition},
  pages={1290--1299},
  year={2022}
}

@inproceedings{assran2022masked,
  title={Masked siamese networks for label-efficient learning},
  author={Assran, Mahmoud and Caron, Mathilde and Misra, Ishan and Bojanowski, Piotr and Bordes, Florian and Vincent, Pascal and Joulin, Armand and Rabbat, Mike and Ballas, Nicolas},
  booktitle={European conference on computer vision},
  pages={456--473},
  year={2022},
  organization={Springer}
}

@article{cheng2021per,
  title={Per-pixel classification is not all you need for semantic segmentation},
  author={Cheng, Bowen and Schwing, Alex and Kirillov, Alexander},
  journal={Advances in neural information processing systems},
  volume={34},
  pages={17864--17875},
  year={2021}
}

@article{grill2020bootstrap,
  title={Bootstrap your own latent-a new approach to self-supervised learning},
  author={Grill, Jean-Bastien and Strub, Florian and Altch{\'e}, Florent and Tallec, Corentin and Richemond, Pierre and Buchatskaya, Elena and Doersch, Carl and Avila Pires, Bernardo and Guo, Zhaohan and Gheshlaghi Azar, Mohammad and others},
  journal={Advances in neural information processing systems},
  volume={33},
  pages={21271--21284},
  year={2020}
}

@inproceedings{chen2021empirical,
  title={An empirical study of training self-supervised vision transformers},
  author={Chen, Xinlei and Xie, Saining and He, Kaiming},
  booktitle={Proceedings of the IEEE/CVF international conference on computer vision},
  pages={9640--9649},
  year={2021}
}

@inproceedings{tao2023siamese,
  title={Siamese image modeling for self-supervised vision representation learning},
  author={Tao, Chenxin and Zhu, Xizhou and Su, Weijie and Huang, Gao and Li, Bin and Zhou, Jie and Qiao, Yu and Wang, Xiaogang and Dai, Jifeng},
  booktitle={Proceedings of the IEEE/CVF Conference on Computer Vision and Pattern Recognition},
  pages={2132--2141},
  year={2023}
}

@article{li2022semmae,
  title={Semmae: Semantic-guided masking for learning masked autoencoders},
  author={Li, Gang and Zheng, Heliang and Liu, Daqing and Wang, Chaoyue and Su, Bing and Zheng, Changwen},
  journal={Advances in Neural Information Processing Systems},
  volume={35},
  pages={14290--14302},
  year={2022}
}

@article{loshchilov2017decoupled,
  title={Decoupled weight decay regularization},
  author={Loshchilov, I},
  journal={arXiv preprint arXiv:1711.05101},
  year={2017}
}

@misc{sirinukunwattana2016gland,
      title={Gland Segmentation in Colon Histology Images: The GlaS Challenge Contest}, 
      author={Korsuk Sirinukunwattana and Josien P. W. Pluim and Hao Chen and Xiaojuan Qi and Pheng-Ann Heng and Yun Bo Guo and Li Yang Wang and Bogdan J. Matuszewski and Elia Bruni and Urko Sanchez and Anton Böhm and Olaf Ronneberger and Bassem Ben Cheikh and Daniel Racoceanu and Philipp Kainz and Michael Pfeiffer and Martin Urschler and David R. J. Snead and Nasir M. Rajpoot},
      year={2016},
      eprint={1603.00275},
      archivePrefix={arXiv},
      primaryClass={cs.CV}
}

@inproceedings{xie2022simmim,
  title={Simmim: A simple framework for masked image modeling},
  author={Xie, Zhenda and Zhang, Zheng and Cao, Yue and Lin, Yutong and Bao, Jianmin and Yao, Zhuliang and Dai, Qi and Hu, Han},
  booktitle={Proceedings of the IEEE/CVF conference on computer vision and pattern recognition},
  pages={9653--9663},
  year={2022}
}

@article{darcet2025cluster,
  title={Cluster and predict latent patches for improved masked image modeling},
  author={Darcet, Timoth{\'e}e and Baldassarre, Federico and Oquab, Maxime and Mairal, Julien and Bojanowski, Piotr},
  journal={arXiv preprint arXiv:2502.08769},
  year={2025}
}

@article{sablayrolles2018spreading,
  title={Spreading vectors for similarity search},
  author={Sablayrolles, Alexandre and Douze, Matthijs and Schmid, Cordelia and J{\'e}gou, Herv{\'e}},
  journal={arXiv preprint arXiv:1806.03198},
  year={2018}
}

@article{hu2022lora,
  title={Lora: Low-rank adaptation of large language models.},
  author={Hu, Edward J and Shen, Yelong and Wallis, Phillip and Allen-Zhu, Zeyuan and Li, Yuanzhi and Wang, Shean and Wang, Lu and Chen, Weizhu and others},
  journal={ICLR},
  volume={1},
  number={2},
  pages={3},
  year={2022}
}

@article{zhou2017places,
  title={Places: A 10 million image database for scene recognition},
  author={Zhou, Bolei and Lapedriza, Agata and Khosla, Aditya and Oliva, Aude and Torralba, Antonio},
  journal={IEEE transactions on pattern analysis and machine intelligence},
  volume={40},
  number={6},
  pages={1452--1464},
  year={2017},
  publisher={IEEE}
}

@inproceedings{radford2021learning,
  title={Learning transferable visual models from natural language supervision},
  author={Radford, Alec and Kim, Jong Wook and Hallacy, Chris and Ramesh, Aditya and Goh, Gabriel and Agarwal, Sandhini and Sastry, Girish and Askell, Amanda and Mishkin, Pamela and Clark, Jack and others},
  booktitle={International conference on machine learning},
  pages={8748--8763},
  year={2021},
  organization={PmLR}
}

@inproceedings{zhang2025assessing,
  title={Assessing and Learning Alignment of Unimodal Vision and Language Models},
  author={Zhang, Le and Yang, Qian and Agrawal, Aishwarya},
  booktitle={Proceedings of the Computer Vision and Pattern Recognition Conference},
  pages={14604--14614},
  year={2025}
}

@article{mairal2008supervised,
  title={Supervised dictionary learning},
  author={Mairal, Julien and Ponce, Jean and Sapiro, Guillermo and Zisserman, Andrew and Bach, Francis},
  journal={Advances in neural information processing systems},
  volume={21},
  year={2008}
}

@String(ICLR = {Int. Conf. Learn. Represent.})

@String(ICLR  = {ICLR})

@inproceedings{carion2020end,
  title={End-to-end object detection with transformers},
  author={Carion, Nicolas and Massa, Francisco and Synnaeve, Gabriel and Usunier, Nicolas and Kirillov, Alexander and Zagoruyko, Sergey},
  booktitle={European conference on computer vision},
  pages={213--229},
  year={2020},
  organization={Springer}
}

@inproceedings{li2023blip,
  title={Blip-2: Bootstrapping language-image pre-training with frozen image encoders and large language models},
  author={Li, Junnan and Li, Dongxu and Savarese, Silvio and Hoi, Steven},
  booktitle={International conference on machine learning},
  pages={19730--19742},
  year={2023},
  organization={PMLR}
}

@article{yu2024image,
  title={An image is worth 32 tokens for reconstruction and generation},
  author={Yu, Qihang and Weber, Mark and Deng, Xueqing and Shen, Xiaohui and Cremers, Daniel and Chen, Liang-Chieh},
  journal={Advances in Neural Information Processing Systems},
  volume={37},
  pages={128940--128966},
  year={2024}
}

@inproceedings{caron2021emerging,
  title={Emerging properties in self-supervised vision transformers},
  author={Caron, Mathilde and Touvron, Hugo and Misra, Ishan and J{\'e}gou, Herv{\'e} and Mairal, Julien and Bojanowski, Piotr and Joulin, Armand},
  booktitle={Proceedings of the IEEE/CVF international conference on computer vision},
  pages={9650--9660},
  year={2021}
}

@article{oquab2023dinov2,
  title={Dinov2: Learning robust visual features without supervision},
  author={Oquab, Maxime and Darcet, Timoth{\'e}e and Moutakanni, Th{\'e}o and Vo, Huy and Szafraniec, Marc and Khalidov, Vasil and Fernandez, Pierre and Haziza, Daniel and Massa, Francisco and El-Nouby, Alaaeldin and others},
  journal={arXiv preprint arXiv:2304.07193},
  year={2023}
}

@article{ding2008convex,
  title={Convex and semi-nonnegative matrix factorizations},
  author={Ding, Chris HQ and Li, Tao and Jordan, Michael I},
  journal={IEEE transactions on pattern analysis and machine intelligence},
  volume={32},
  number={1},
  pages={45--55},
  year={2008},
  publisher={IEEE}
}

@inproceedings{deng2009imagenet,
  title={Imagenet: A large-scale hierarchical image database},
  author={Deng, Jia and Dong, Wei and Socher, Richard and Li, Li-Jia and Li, Kai and Fei-Fei, Li},
  booktitle={2009 IEEE conference on computer vision and pattern recognition},
  pages={248--255},
  year={2009},
  organization={Ieee}
}

@inproceedings{chang2022maskgit,
  title={Maskgit: Masked generative image transformer},
  author={Chang, Huiwen and Zhang, Han and Jiang, Lu and Liu, Ce and Freeman, William T},
  booktitle={Proceedings of the IEEE/CVF conference on computer vision and pattern recognition},
  pages={11315--11325},
  year={2022}
}

@inproceedings{esser2021taming,
  title={Taming transformers for high-resolution image synthesis},
  author={Esser, Patrick and Rombach, Robin and Ommer, Bjorn},
  booktitle={Proceedings of the IEEE/CVF conference on computer vision and pattern recognition},
  pages={12873--12883},
  year={2021}
}

@inproceedings{NIPS2017_8a1d6947,
 author = {Heusel, Martin and Ramsauer, Hubert and Unterthiner, Thomas and Nessler, Bernhard and Hochreiter, Sepp},
 booktitle = {Advances in Neural Information Processing Systems},
 editor = {I. Guyon and U. Von Luxburg and S. Bengio and H. Wallach and R. Fergus and S. Vishwanathan and R. Garnett},
 pages = {},
 publisher = {Curran Associates, Inc.},
 title = {GANs Trained by a Two Time-Scale Update Rule Converge to a Local Nash Equilibrium},
 url = {https://proceedings.neurips.cc/paper_files/paper/2017/file/8a1d694707eb0fefe65871369074926d-Paper.pdf},
 volume = {30},
 year = {2017}
}

@inproceedings{cordts2016cityscapes,
  title={The cityscapes dataset for semantic urban scene understanding},
  author={Cordts, Marius and Omran, Mohamed and Ramos, Sebastian and Rehfeld, Timo and Enzweiler, Markus and Benenson, Rodrigo and Franke, Uwe and Roth, Stefan and Schiele, Bernt},
  booktitle={Proceedings of the IEEE conference on computer vision and pattern recognition},
  pages={3213--3223},
  year={2016}
}

@article{everingham2010pascal,
  title={The pascal visual object classes (voc) challenge},
  author={Everingham, Mark and Van Gool, Luc and Williams, Christopher KI and Winn, John and Zisserman, Andrew},
  journal={International journal of computer vision},
  volume={88},
  number={2},
  pages={303--338},
  year={2010},
  publisher={Springer}
}

@article{zhou2021ibot,
  title={ibot: Image bert pre-training with online tokenizer},
  author={Zhou, Jinghao and Wei, Chen and Wang, Huiyu and Shen, Wei and Xie, Cihang and Yuille, Alan and Kong, Tao},
  journal={arXiv preprint arXiv:2111.07832},
  year={2021}
}

@inproceedings{parkhi2012cats,
  title={Cats and dogs},
  author={Parkhi, Omkar M and Vedaldi, Andrea and Zisserman, Andrew and Jawahar, CV},
  booktitle={2012 IEEE conference on computer vision and pattern recognition},
  pages={3498--3505},
  year={2012},
  organization={IEEE}
}

@inproceedings{bossard14,
  title = {Food-101 -- Mining Discriminative Components with Random Forests},
  author = {Bossard, Lukas and Guillaumin, Matthieu and Van Gool, Luc},
  booktitle = {European Conference on Computer Vision},
  year = {2014}
}

@inproceedings{zhou2017scene,
  title={Scene parsing through ade20k dataset},
  author={Zhou, Bolei and Zhao, Hang and Puig, Xavier and Fidler, Sanja and Barriuso, Adela and Torralba, Antonio},
  booktitle={Proceedings of the IEEE conference on computer vision and pattern recognition},
  pages={633--641},
  year={2017}
}

@article{el2024scalable,
  title={Scalable pre-training of large autoregressive image models},
  author={El-Nouby, Alaaeldin and Klein, Michal and Zhai, Shuangfei and Bautista, Miguel Angel and Toshev, Alexander and Shankar, Vaishaal and Susskind, Joshua M and Joulin, Armand},
  journal={arXiv preprint arXiv:2401.08541},
  year={2024}
}

@article{caron2020unsupervised,
  title={Unsupervised learning of visual features by contrasting cluster assignments},
  author={Caron, Mathilde and Misra, Ishan and Mairal, Julien and Goyal, Priya and Bojanowski, Piotr and Joulin, Armand},
  journal={Advances in neural information processing systems},
  volume={33},
  pages={9912--9924},
  year={2020}
}

@article{zhang2022improving,
  title={Improving vae-based representation learning},
  author={Zhang, Mingtian and Xiao, Tim Z and Paige, Brooks and Barber, David},
  journal={arXiv preprint arXiv:2205.14539},
  year={2022}
}

@article{chen2024deconstructing,
  title={Deconstructing denoising diffusion models for self-supervised learning},
  author={Chen, Xinlei and Liu, Zhuang and Xie, Saining and He, Kaiming},
  journal={arXiv preprint arXiv:2401.14404},
  year={2024}
}

@inproceedings{assran2023self,
  title={Self-supervised learning from images with a joint-embedding predictive architecture},
  author={Assran, Mahmoud and Duval, Quentin and Misra, Ishan and Bojanowski, Piotr and Vincent, Pascal and Rabbat, Michael and LeCun, Yann and Ballas, Nicolas},
  booktitle={Proceedings of the IEEE/CVF Conference on Computer Vision and Pattern Recognition},
  pages={15619--15629},
  year={2023}
}

@inproceedings{baevski2022data2vec,
  title={Data2vec: A general framework for self-supervised learning in speech, vision and language},
  author={Baevski, Alexei and Hsu, Wei-Ning and Xu, Qiantong and Babu, Arun and Gu, Jiatao and Auli, Michael},
  booktitle={International conference on machine learning},
  pages={1298--1312},
  year={2022},
  organization={PMLR}
}

@inproceedings{zhang2018unreasonable,
  title={The unreasonable effectiveness of deep features as a perceptual metric},
  author={Zhang, Richard and Isola, Phillip and Efros, Alexei A and Shechtman, Eli and Wang, Oliver},
  booktitle={Proceedings of the IEEE conference on computer vision and pattern recognition},
  pages={586--595},
  year={2018}
}

@inproceedings{devlin2019bert,
  title={Bert: Pre-training of deep bidirectional transformers for language understanding},
  author={Devlin, Jacob and Chang, Ming-Wei and Lee, Kenton and Toutanova, Kristina},
  booktitle={Proceedings of the 2019 conference of the North American chapter of the association for computational linguistics: human language technologies, volume 1 (long and short papers)},
  pages={4171--4186},
  year={2019}
}

@article{brown2020language,
  title={Language models are few-shot learners},
  author={Brown, Tom and Mann, Benjamin and Ryder, Nick and Subbiah, Melanie and Kaplan, Jared D and Dhariwal, Prafulla and Neelakantan, Arvind and Shyam, Pranav and Sastry, Girish and Askell, Amanda and others},
  journal={Advances in neural information processing systems},
  volume={33},
  pages={1877--1901},
  year={2020}
}

@article{van2025joint,
  title={Joint Embedding vs Reconstruction: Provable Benefits of Latent Space Prediction for Self Supervised Learning},
  author={Van Assel, Hugues and Ibrahim, Mark and Biancalani, Tommaso and Regev, Aviv and Balestriero, Randall},
  journal={arXiv preprint arXiv:2505.12477},
  year={2025}
}
\bibliographystyle{icml2026}

%%%%%%%%%%%%%%%%%%%%%%%%%%%%%%%%%%%%%%%%%%%%%%%%%%%%%%%%%%%%%%%%%%%%%%%%%%%%%%%
%%%%%%%%%%%%%%%%%%%%%%%%%%%%%%%%%%%%%%%%%%%%%%%%%%%%%%%%%%%%%%%%%%%%%%%%%%%%%%%
% APPENDIX
%%%%%%%%%%%%%%%%%%%%%%%%%%%%%%%%%%%%%%%%%%%%%%%%%%%%%%%%%%%%%%%%%%%%%%%%%%%%%%%
%%%%%%%%%%%%%%%%%%%%%%%%%%%%%%%%%%%%%%%%%%%%%%%%%%%%%%%%%%%%%%%%%%%%%%%%%%%%%%%
\newpage
\appendix
\onecolumn
\appendix
% \section{Statements}

% \subsection{Ethics Statement}
% This work adheres to the ICLR Code of Ethics. Our research is based on publicly available datasets (e.g., ImageNet-1K, ADE20K, Cityscapes, Pascal VOC) and does not involve human subjects, private data, or personally identifiable information. We follow standard licensing terms for all datasets used. The proposed framework, STELLAR, is a general-purpose method for self-supervised representation learning and is not designed for harmful or sensitive applications. 

% \subsection{Reproducibility Statement}
% We have made significant efforts to ensure the reproducibility of our results. Detailed descriptions of the STELLAR framework, training objectives, and experimental setups are provided in the main text and appendix. We report all datasets used, including preprocessing steps and evaluation protocols, and we describe ablation studies to clarify the contribution of each component. We will release the source code and pretrained model checkpoints after the internal review process to further facilitate reproducibility.

% \subsection{Use of Large Language Models}
% Large language models (ChatGPT) were used exclusively for language refinement, including polishing grammar, phrasing, and clarity of the manuscript. They were not used for research ideation, methodological design, experimental implementation, data analysis, or drawing conclusions. All scientific contributions of this work are entirely by the authors.

\section{Implementation Details}
\subsection{STELLAR Training}
We trained STELLAR with ViT models at size base, large, and huge, along with the latent queries, projection layers, clustering head, and a 6-layer ViT decoder. In the default setting, we initialized the ViT part in the encoder from public MAE checkpoint, and trained for 150 epochs for STELLAR-B, 100 epochs for STELLAR-L, and 50 epochs for STELLAR-H. We used 16 NVIDIA A100-80GB with batch size 128 each, totaling 2048. We used AdamW\citep{loshchilov2017decoupled} with base learning rate $1.5\times10^{-4}$ for STELLAR-B, and $5\times10^{-5}$ for STELLAR-L and STELLAR-H.

For concept clustering, we used 16384 prototypes for sparse and CLS tokens each. The projector is a 2-layer MLP before the prototype layer. We used 3 steps of Sinkhon-Knopp algorithm. The temperature in sparse-dense cosine similarity softmax is 0.06. We used 6-8 random masked views to align the sparse tokens, and additional 6-8 local crops to align the CLS token. Global views are of random scale 36\% to 100\%, and local view are of random scale 6\% to 36\%. We also apply color jittering, grascaling and Gaussian blurring.

In the ablation study of random prior, we trained the model from scratch and used exponential moving average (EMA) updated momentum encoder to encode the target prototype assignments in the warm-up stage. We EMA updated the full encoder (ViT, latent queries, projection, clustering head with momentum 0.996. The momentum encoder was used to encode a global view of the image into target prototype assignments, for both clustering loss and alignment loss. The masking ratio was 0.6 in the warm-up stage, and 0.8 during standard training. We trained the model with 150 epochs of EMA warm-up and 75 epochs of standard training.

\subsection{Evaluation Protocol}\label{eval}
For STELLAR and all baseline models, we evaluated the frozen feature from the pretrained model with linear probing. We used layer norm in classification tasks, and batch norm in segmentation tasks, followed by a single linear layer predicting the class of the image or patch. For all benchmarks, we split 10\% from the training set for validation. We tuned hyper-parameter with learning rate $1\times 10^{-5}, 2\times 10^{-5}, 5\times 10^{-5}, 1\times 10^{-4}, 2\times 10^{-4}, 5\times 10^{-4}, 1\times 10^{-3}, 2\times 10^{-3}, 5\times 10^{-3}, 1\times 10^{-2}$, and batch size 64, 128, 256, 512, 1024, 2048, 4096, 8192.

As the SSL methods varies across different baseline models, for classification tasks we used the mean-pooled feature from the representations where the corresponding SSL method was performed, e.g. the global CLS token for DINO, and dense patch tokens for MAE. We noticed the linear probing accuracy can vary depending on the pooling choice, and conducted experiments by using different types of tokens for each model, with results in Table \ref{pooling-table}. We observed that the SSL-ed are typically the best choice for linear probing, except for iBOT, which highly relies on the global CLS token for classification, even though the model was trained with MIM. In contrast, STELLAR and MAE are relatively more robust to token choices.

\begin{table}[hb]
\caption{ImageNet-1K linear probing accuracy (\%) by pooling different tokens. We mark in \textbf{bold} the tokens on which the specific SSL method was applied, and the top accuracy for each method.}
\label{pooling-table}
\begin{center}
\begin{tabular}{c|cc|cc|ccc|cc}
\multicolumn{1}{c}{}   & \multicolumn{2}{c}{DINO} &  \multicolumn{2}{c}{MAE} & \multicolumn{3}{c}{iBOT} &\multicolumn{2}{c}{STELLAR (ours)} \\
\hline 
tokens &  \textbf{global} & dense & global & \textbf{dense} & global & dense & \textbf{gl.+de.} & \textbf{sparse} &  dense \\
\hline 
lin. acc.  & \textbf{76.46} & 70.31 & 65.61 & \textbf{66.32} & \textbf{76.40} & 71.44 &  71.58 & \textbf{73.26} &  72.21
\end{tabular}
\end{center}
\end{table}

\begin{table}[hb]
\caption{List of baseline models and SSL method type.}
\label{tab:baselines}
\begin{center}
\begin{tabular}{l l c c c }
\multicolumn{1}{c}{\bf Model} &
\multicolumn{1}{c}{Reference} &
\multicolumn{1}{c}{Method} &
\multicolumn{1}{c}{SSL space} &
\multicolumn{1}{c}{SSL tokens}
\\ \hline 
BYOL & \cite{grill2020bootstrap} & augmentation alignment  &  latent & global \\
MoCo v3 & \cite{chen2021empirical} & contrastive learning  &  latent & global  \\
DINO & \cite{caron2021emerging} & augmentation alignment   &  latent & global \\
MSN  & \cite{assran2022masked}  & masked alignment  &  latent & global  \\
DenseCL & \cite{wang2021dense} & contrastive learning &  latent & dense \\
Data2Vec & \cite{baevski2022data2vec}  & latent MIM   & latent  & dense \\
SiameseIM & \cite{tao2023siamese}  &  latent MIM  & latent  & dense   \\
IJEPA & \cite{assran2023self}  &  latent MIM & latent  & dense   \\
iBOT & \cite{zhou2021ibot}  &  align + latent MIM  & latent  & global+dense  \\
BEIT & \cite{bao2021beit}  & token MIM  &  image  & dense \\
SimMIM  & \cite{xie2022simmim} & pixel MIM  & image  & dense  \\
MAE  & \cite{he2022masked}  & pixel MIM  &  image  & dense  \\
SemMAE  & \cite{li2022semmae}  & pixel MIM  &  image  & dense \\
TiTok  & \cite{yu2024image}  & reconstruction + clustering  &  image  & sparse  \\
AIM      & \cite{el2024scalable}  & autoregressive  & image & dense \\
DINOv2      & \cite{oquab2023dinov2}  & align + latent MIM   & latent & global+dense
\end{tabular}
\end{center}
\end{table}

\section{Additional Results}\label{app:results}

\subsection{Effect of pretraining data} 
We pretrained separate STELLAR versions on ImageNet-1K, Places365~\citep{zhou2017places} and compared their linear probing performance in Table \ref{data-table}.
\begin{table}[ht]
\caption{Effect of pretraining data.}
\label{data-table}
\begin{center}
\begin{tabular}{c|cc}
& \multicolumn{2}{c}{linear probing acc.} \\
 Pretraining data & ImageNet-1K & Places 365 \\
\hline 
ImageNet-1K & 76.94 & 49.25 \\
Places365 & 66.08 & 51.98
\end{tabular}
\end{center}
\end{table}

\subsection{Semantics from different features} 
We conducted linear probing of different mean-pooled features of different types, and compared in Table \ref{feat-table}. Sparse feature showed strongest global understanding quality.
\begin{table}[ht]
\caption{Semantics in different features}
\label{feat-table}
\begin{center}
\begin{tabular}{c|ccc}
 Feature & sparse & cls & dense \\
\hline 
IN-1K lin. acc (\%) & 73.26 & 72.23 & 72.21 
\end{tabular}
\end{center}
\end{table}

\subsection{Concept alignment with language}
Inspired by \cite{zhang2025assessing}, we used frozen feature from STELLAR and aligned with the text tower of CLIP~\citep{radford2021learning} with a single attention pooled probing layer. The evaluation on vision language tasks with comparison to baseline models are shown in Table \ref{vlm-table}.

\begin{table}[ht]
\caption{Language alignment evaluation.}
\label{vlm-table}
\begin{center}
\begin{tabular}{c|cc|cc|cc|c}
\multicolumn{1}{c}{}   & \multicolumn{2}{c}{IN-1K 0-shot} &  \multicolumn{2}{c}{MS COCO} & \multicolumn{2}{c}{Winoground} & MMVP \\
\hline 
  &      @1  & @5 & T2I & I2T & Text & Image & Avg. \\
\hline 
MAE &   23.18 & 50.43 & 11.28 & 13.46 & 20.75 & 9.00 & 19.26 \\
iBOT &  50.01 & 80.43 & 20.79 & 29.38 & 24.75 & 12.00 & 18.52 \\
STELLAR &   51.53 & 80.04 & 17.94 & 22.34 & 26.25 & 8.25 & 19.26 \\
\hline
CLIP  &  72.7 & - & 43.0 & 59.7 & 30.5 & 11.5 & 20.0
\end{tabular}
\end{center}
\end{table}

\subsection{Finetuning}
We performed finetuning for STELLAR on ImageNet-1K classification and ADE20K segmentation, and compared with baseline models. We used the same evaluation protocol as in Sec. \ref{eval}, with the backbone unfrozen and finetuned for 75 epochs. We used ViT-B for all models. The finetuning results are shown in Table \ref{finetune-table}. STELLAR showed consistent performance gain across different tasks, and close to the top model iBOT with slight difference.

\begin{table}[ht]

\caption{Finetuning performance in ImageNet-1K classification accuracy and ADE20K segmentation mIOU (\%). We show in parentheses the gain over the respective linear probing results.}
\label{finetune-table}
\begin{center}
\begin{tabular}{c|ccc}
 Model & ImageNet-1K Acc. & ADE20K mIOU \\
\hline 
DINO & 79.58 (+3.12) & 39.22 (+12.35) \\
MAE & 77.75 (+11.43) & 40.33 (+9.42) \\
iBOT & 80.72 (+9.14) & 42.76 (+10.97) \\
\hline
STELLAR & 80.05 (+6.78) & 41.98 (+10.65)
\end{tabular}
\end{center}
\end{table}

\subsection{Efficiency analysis}
To analyze the efficiency of the STELLAR framework, we printed the processing time of the main components in the STELLAR framework with one A100 GPU at different batch sizes. Encoding the main global view of the image takes up most of the processing time, followed by encoding the masked views (8 views at 80\% masking ratio) and decoding to the original image. The Sinkhorn-Knopp algorithm used for clustering and the Sinkhorn algorithm used in optimal transport matching take up much less amount of time, and their total processing time stay at similar level when increasing the batch size. 

In comparison to the Sinkhorn matching algorithm we used in our experiments, we show the processing time using an alternative Hungarian matching algorithm commonly used in previous literature such as Sparse R-CNN~\citep{sun2021sparse}, DETR~\citep{carion2020end} and MaskFormer~\citep{cheng2021per}. As the implementation of the exact matching is not scalable with GPU parallelization, it's computational time increases linearly with the batch size. At batch size 64, it is already 6 times of the encoder processing, while the Sinkhorn algorithm is over 100 times faster. For this reason, we added a small entropy regularization term in the bipartite matching objective, allowing us to use the Sinkhorn algorithm for efficient matching with GPU parallelization.

\begin{table}[ht]
\caption{Processing time (s) of the main components in the STELLAR framework with one A100 GPU at different batch sizes. In comparison to the Sinkhorn matching algorithm we used in our experiments, we show the processing time using an alternative Hungarian matching algorithm commonly used in previous literature (shown in gray).}
\label{finetune-table}
\begin{center}
\begin{tabular}{c|ccccc}
 Batch size & 4 & 8 & 16 & 32 & 64  \\
\hline 
Encoder & 8.2 $\times 10^{-3}$ & 9.1 $\times 10^{-3}$ & 1.4 $\times 10^{-2}$ & 2.0 $\times 10^{-2}$ & 3.2 $\times 10^{-2}$ \\
Decoder & $4.6 \times 10^{-3}$ & 6.8 $\times 10^{-3}$ & 8.8 $\times 10^{-3}$ & 1.2 $\times 10^{-2}$ & 1.5 $\times 10^{-2}$ \\
Mask encoding & $7.9 \times 10^{-3}$ & 8.9 $\times 10^{-3}$ & 1.1 $\times 10^{-2}$ & 1.8 $\times 10^{-2}$ & 1.7 $\times 10^{-2}$\\
SK clustering & $3.4 \times 10^{-4}$ & 3.4 $\times 10^{-4}$ & 3.4 $\times 10^{-4}$ & $3.7 \times 10^{-4}$ & 3.9 $\times 10^{-4}$  \\
Sinkhorn matching & $1.4 \times 10^{-3}$ & 1.4 $\times 10^{-3}$ & 1.4 $\times 10^{-3}$ & $1.4 \times 10^{-3}$ & 1.2 $\times 10^{-3}$ \\
\hline
\color{gray}{Hungarian matching} & \color{gray}{$5.7 \times 10^{-3}$} & \color{gray}{1.7 $\times 10^{-2}$} & \color{gray}{4.0 $\times 10^{-2}$} & \color{gray}{9.0 $\times 10^{-2}$} & \color{gray}{1.8 $\times 10^{-1}$}
\end{tabular}
\end{center}
\end{table}

% \begin{table}[t]
% \caption{Model performance with different configurations}
% \label{mse-table}
% \begin{center}
% \begin{tabular}{cccccc|cc}
% \# tk.  & \multicolumn{1}{c}{Recon.}  &\multicolumn{1}{c}{Cluster} &  Set align & CLS align & KoLeo & IN1k lin. & ADE lin. \\
% \hline \\
% 8 &    code  & \checkmark & \checkmark & \checkmark & \checkmark & 72.97 & 31.04 \\
% 16  &   code    & \checkmark  & \checkmark & \checkmark & \checkmark & 73.26 & 31.33\\
% \hline 
% \multicolumn{8}{l}{\it Ablation models versions} \\
% 8 &    pixel  & \checkmark & \checkmark & \checkmark & \checkmark & 72.16 (-0.81) & 30.04 (-1.00) \\
% 16  &  \ding{55}   & \checkmark  & \checkmark & \checkmark & \checkmark & 72.44 (-0.82) & 29.94 (-1.39) \\
% 16  &   code    &  \ding{55}   &  \ding{55}  &  \ding{55}  & \checkmark & 52.07 (-21.19) & 20.46 (-10.87) \\
% 16  &   code    & \checkmark  & \ding{55}  & \ding{55}  & \checkmark & 2.73 (-70.53) & 1.93 (-29.39) \\
% 16  &   code    & \checkmark  & \checkmark & \ding{55}  & \checkmark & 70.79 (-2.47) & 30.20 (-1.12) \\
% 16  &   code    & \checkmark  & \checkmark & \checkmark & \ding{55}  & 72.05 (-1.21) & 30.10 (-1.23) \\
% \end{tabular}
% \end{center}
% \end{table}

\section{Additional Illustration}
\begin{figure}
    \centering
    \includegraphics[trim={1cm 13.5cm 7cm 1cm},clip, width=\linewidth]{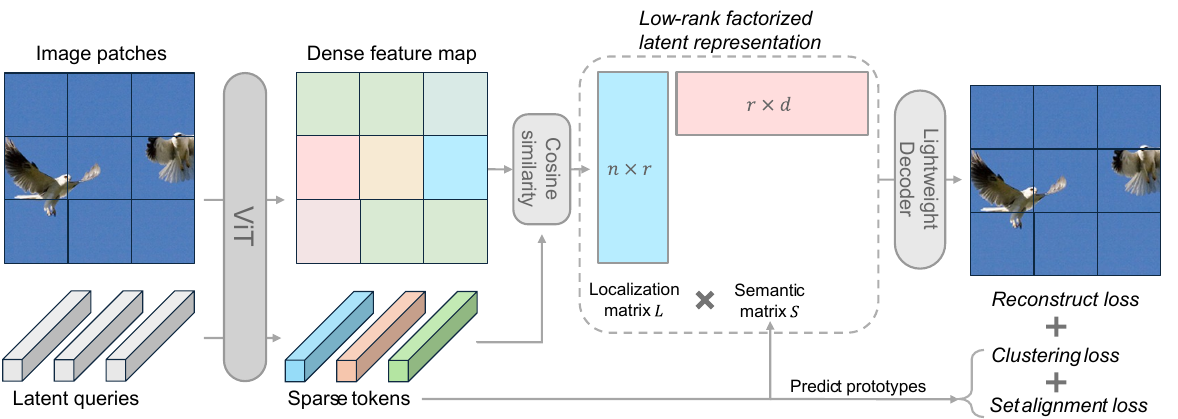}
    \caption{\small The STELLAR framework. We use a vanilla ViT to extract sparse tokens from an image, and model the latent representation as a low-rank matrix factorization, ensuring reconstruction of the original image. Clustering loss and set alignment loss are applied on the disentangled sparse tokens.}
    \label{fig:stellar}
\end{figure}

\begin{figure}
    \centering
    \includegraphics[trim={1cm 10.5cm 1cm 1cm},clip, width=\linewidth]{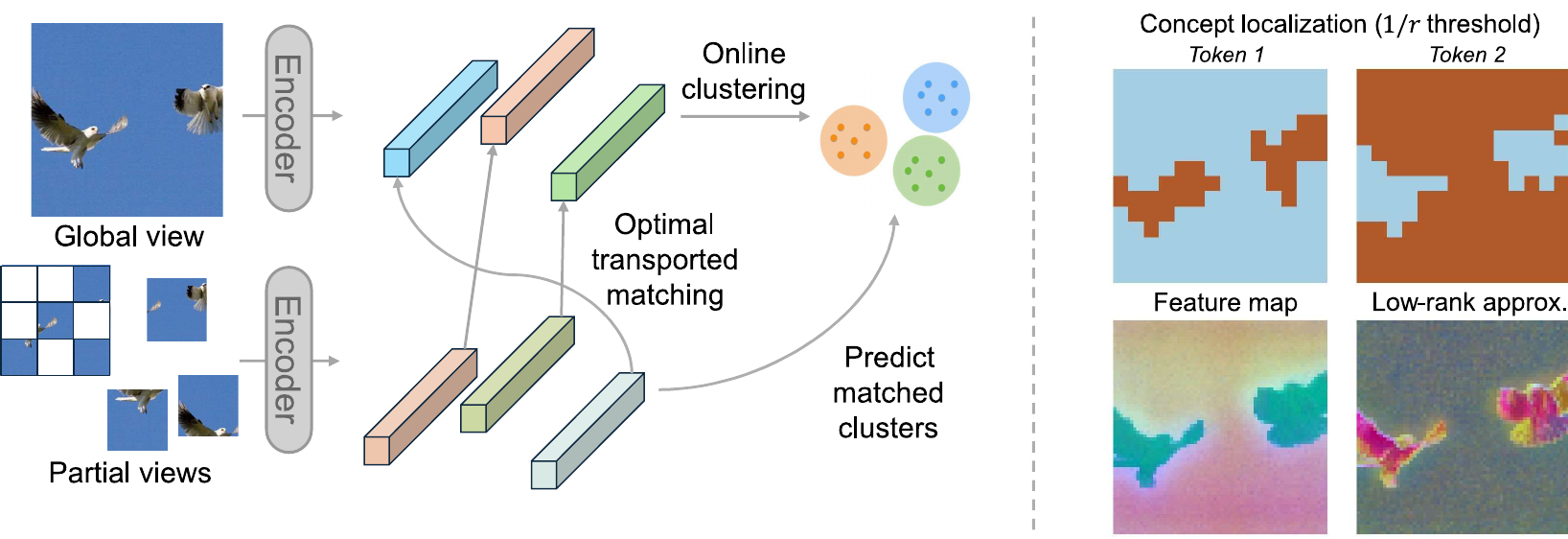}
    \caption{\small Left: Concept clustering and alignment workflow. Right: visualization of learned representation.}
    \label{fig:align}
\end{figure}

See Fig.~\ref{fig:stellar} and \ref{fig:align}.
%%%%%%%%%%%%%%%%%%%%%%%%%%%%%%%%%%%%%%%%%%%%%%%%%%%%%%%%%%%%%%%%%%%%%%%%%%%%%%%
%%%%%%%%%%%%%%%%%%%%%%%%%%%%%%%%%%%%%%%%%%%%%%%%%%%%%%%%%%%%%%%%%%%%%%%%%%%%%%%

\end{document}